\documentclass[11pt]{article}

\usepackage[a4paper,margin=1in]{geometry}
\usepackage[T1]{fontenc}
\usepackage[utf8]{inputenc}
\usepackage{lmodern}
\usepackage{microtype}
\usepackage{amsmath,amssymb,mathtools}
\usepackage{booktabs}
\usepackage{longtable}
\usepackage{tabularx}
\usepackage{array}
\usepackage{multirow}
\usepackage{graphicx}
\usepackage{xcolor}
\usepackage{enumitem}
\usepackage{listings}
\usepackage{caption}
\usepackage{subcaption}
\usepackage{float}
\usepackage{placeins}
\usepackage{tikz}
\usetikzlibrary{arrows.meta,positioning,fit,shapes.geometric}
\usepackage{hyperref}
\usepackage{url}
\usepackage{fancyhdr}

\definecolor{CeloInk}{HTML}{202124}
\definecolor{CeloGold}{HTML}{B78B2E}
\definecolor{CeloBlue}{HTML}{2A6F97}
\definecolor{CeloGreen}{HTML}{2A7F62}
\definecolor{CeloRed}{HTML}{A33F3F}
\definecolor{CeloLight}{HTML}{F7F3E8}
\definecolor{CeloGray}{HTML}{5F6368}

\hypersetup{
  colorlinks=true,
  linkcolor=CeloBlue,
  citecolor=CeloGreen,
  urlcolor=CeloBlue,
  pdftitle={From a Multilingual Streaming ASR Backbone to Kenyan-Language Systems: Data-Centric Adaptation of Nemotron 3.5 for Kikuyu, Dholuo, and Kalenjin},
  pdfauthor={Mark Gatere}
}

\setlist[itemize]{leftmargin=1.5em,itemsep=0.25em,topsep=0.35em}
\setlist[enumerate]{leftmargin=1.7em,itemsep=0.25em,topsep=0.35em}

\newcolumntype{Y}{>{\raggedright\arraybackslash}X}
\newcolumntype{C}[1]{>{\centering\arraybackslash}p{#1}}
\newcolumntype{L}[1]{>{\raggedright\arraybackslash}p{#1}}

\lstdefinestyle{paperlisting}{
  basicstyle=\ttfamily\small,
  frame=single,
  rulecolor=\color{black!18},
  backgroundcolor=\color{black!2},
  breaklines=true,
  columns=fullflexible,
  keepspaces=true,
  showstringspaces=false,
  xleftmargin=0.3em,
  xrightmargin=0.3em
}
\lstdefinelanguage{json}{
  basicstyle=\ttfamily\small,
  string=[s]{"}{"},
  stringstyle=\color{CeloGreen},
  comment=[l]{//},
  commentstyle=\color{CeloGray}\itshape,
  morecomment=[s]{/*}{*/},
  keywords={true,false,null},
  keywordstyle=\color{CeloBlue}\bfseries,
  literate=
   *{:}{{{\color{CeloInk}{:}}}}{1}
    {,}{{{\color{CeloInk}{,}}}}{1}
    {\{}{{{\color{CeloInk}{\{}}}}{1}
    {\}}{{{\color{CeloInk}{\}}}}}{1}
    {[}{{{\color{CeloInk}{[}}}}{1}
    {]}{{{\color{CeloInk}{]}}}}{1}
}

\newcommand{\wer}{\mathrm{WER}}
\newcommand{\cer}{\mathrm{CER}}
\newcommand{\nscer}{\mathrm{NS\mbox{-}CER}}
\newcommand{\project}{C-elo ASR}
\newcommand{\model}{Nemotron 3.5 ASR Streaming 0.6B}

\title{\textbf{From a Multilingual Streaming ASR Backbone to Kenyan-Language Systems:}\\
Data-Centric Adaptation of Nemotron 3.5 for Kikuyu, Dholuo, and Kalenjin}

\author{
  Mark Gatere\\
  C-elo Labs\\
  \href{mailto:gatere@c-elo.com}{gatere@c-elo.com}\\
\href{https://orcid.org/0009-0009-7616-1824}{ORCID: 0009-0009-7616-1824}
}

\date{July 2026}

\begin{document}

\maketitle

\begin{abstract}
Automatic speech recognition (ASR) for African languages is constrained by orthographic inconsistency, annotation artifacts, missing audio, speaker and domain imbalance, and evaluation procedures that differ from deployment.
We present an end-to-end engineering study adapting NVIDIA Nemotron 3.5 ASR Streaming 0.6B to Kikuyu, Dholuo, and Kalenjin.
Starting from a Kenyan Swahili-adapted checkpoint, we retain its cache-aware FastConformer RNN-T, prompt conditioning, and streaming decoder during full-parameter fine-tuning.
The study covers corpus auditing, Unicode normalization, split checks, duration filtering, low-rate continuation, validation-based checkpoint selection, true-streaming evaluation, artifact preservation, and isolated serving.

On internal, adaptively consulted evaluation sets excluded from gradient updates at context \texttt{[56,13]}, selected Kikuyu and Dholuo models achieve 42.97\% and 33.98\% WER, respectively.
Dholuo records 9.59\% CER and 8.13\% no-space CER under its frozen historical label policy; Kikuyu records 7.79\% no-space CER.
Kalenjin remains a work in progress: v1-v reaches 68.74\% WER on a 2,411-row clean-v3 diagnostic subset excluding annotation markers, digit-bearing references, and targets shorter than three tokens.
Its checkpoint selection used a mixed-source validation manifest containing test-origin rows, so the score is not an independent generalization estimate.
We also report negative findings involving non-speech labels, short-utterance over-generation, boundary-sensitive WER, and cloud job-lifecycle failures.
We make no state-of-the-art claim because the internal sets, repeated consultation, and normalization differ from public benchmarks.
This work provides an auditable account of adapting a multilingual streaming model into language-specific systems without discarding streaming constraints.
\end{abstract}


\vspace{0.8em}
\noindent
\fcolorbox{CeloGold}{CeloLight}{
\begin{minipage}{0.95\textwidth}
\textbf{Status and interpretation.}
Kikuyu and Dholuo are completed engineering case studies with preserved champion checkpoints and live streaming deployments.
Kalenjin is included as an ongoing case study because its failures and successive corpus revisions are scientifically informative.
All numerical results in this report are internal project evaluations unless a public benchmark is named explicitly.
The held-out sets were never used for gradient updates, but repeated milestone inspection influenced research decisions; they therefore must not be represented as a pristine one-shot external benchmark.
\end{minipage}}

\tableofcontents
\newpage

\section{Introduction}

Speech interfaces are increasingly used to search for information, dictate text, operate devices, access public services, and mediate human--computer interaction.
Yet these interfaces remain least reliable for communities whose languages are sparsely represented in commercial and academic training corpora.
The phrase ``low resource'' is often interpreted as a single shortage---too little audio---but operational work reveals a broader problem.
A language may have hundreds of hours of recordings and still be difficult to model because transcriptions encode inconsistent spelling, non-speech events appear as lexical tokens, audio objects are missing, duration tails destabilize batching, domain labels are incomplete, or public splits are not independent in the way their names suggest.

This paper studies that broader problem through the development of \project, a family of streaming ASR systems for Kenyan languages.
The completed systems target Kikuyu (Gikuyu) and Dholuo (Luo); a third system for Kalenjin is under active development.
All three are adapted from \model{} using PyTorch-based NVIDIA NeMo tooling.
Rather than replace the streaming recognizer with an offline encoder--decoder, we retain cache-aware inference from training through deployment.
The resulting research question is therefore not simply whether a large multilingual model can transcribe another language.
It is whether that model can be adapted, evaluated, preserved, and served as a true streaming system under realistic data and infrastructure constraints.

The work is motivated by four observations.
First, a geographically and operationally relevant intermediate checkpoint may provide a practical initialization between a global base model and a Kenyan target language, even when the bridge and target languages are not in the same linguistic subgroup. This study did not isolate its advantage over direct adaptation from the raw NVIDIA base.
Second, data quality decisions are model decisions: replacing, retaining, or dropping an uncertain character sequence changes the supervision seen by every subsequent training stage.
Third, offline WER alone is insufficient for a streaming product because an offline evaluator can conceal state-reset bugs, future-context leakage, unstable partial hypotheses, and latency costs.
Fourth, long fine-tuning runs produce diminishing gains; without a disciplined experiment ledger it is easy to mistake cumulative compute for scientific progress.

\subsection{Research Questions}

The study is organized around the following questions.

\begin{description}[leftmargin=4.0em,style=nextline]
\item[RQ1] What adaptation behavior is observed when a Kenyan Swahili-adapted streaming checkpoint is used to initialize Kikuyu, Dholuo, and Kalenjin models while preserving the original online architecture?
\item[RQ2] Which corpus defects were observed, how were they handled, and how did corpus and normalization revisions change the supervision and measured ASR quality?
\item[RQ3] What improvement trajectory is observed under staged low-rate continuation, and where do returns become small enough to motivate data or decoding work instead?
\item[RQ4] How were trainer validation and true-streaming internal evaluation used, and what limitations arise from repeated held-out consultation and mixed-source selection?
\item[RQ5] Which engineering practices were implemented to move from a research checkpoint to an isolated live service, and which authorization and reproducibility controls remain incomplete?
\end{description}

\subsection{Contributions}

This report makes six contributions.

\begin{enumerate}
\item It documents a complete adaptation path for a cache-aware FastConformer RNN-T from a multilingual base through a Kenyan Swahili bridge to language-specific Kikuyu and Dholuo champions, with an ongoing Kalenjin extension.
\item It provides auditable corpus accounting, including row counts, hours, skip reasons, split provenance, and language-specific cleaning decisions rather than reporting only the nominal size of each source dataset.
\item It reports full continuation lineages and versioned true-streaming evaluations under stated manifests and configurations, including a Kikuyu v4 checkpoint-average candidate that did not outperform the selected checkpoint and diminishing late-stage gains.
\item It analyzes the gap between WER, conventional CER, and no-space CER, showing why word-boundary conventions materially affect interpretation in morphologically rich and orthographically variable settings.
\item It describes a prototype operational architecture for isolated model serving, browser streaming, session-gated configuration, artifact checksums, and private checkpoint mirrors, while identifying authorization controls that remain incomplete.
\item It publishes negative evidence from Kalenjin development---marker contamination, alphanumeric labels, short-utterance behavior, and domain/type stratification---so that future work is not reduced to a single final score.
\end{enumerate}

\subsection{What This Paper Does Not Claim}

The reported numbers do not establish state-of-the-art performance against PazaBench or other public ASR benchmarks and multilingual model families \cite{microsoft2026paza}.
Such a claim requires identical audio, references, normalization, decoding, and metric code.
The internal test sets used here are drawn from the project datasets and were reviewed at multiple milestones.
The results are therefore evidence of progress within a controlled project protocol, not a substitute for an independently administered benchmark.

Likewise, the paper does not claim that the three languages form a homogeneous technical category.
Kikuyu is a Bantu language with contrastive orthographic diacritics; Dholuo and Kalenjin belong to Nilotic language families and present different lexical and segmentation patterns.
The commonality lies in the engineering setting: Kenyan speech data, limited public ASR infrastructure, code switching, heterogeneous domains, and the need for low-latency deployment.

\section{Related Work}

\subsection{End-to-End and Streaming ASR}

End-to-end ASR replaces hand-designed acoustic, pronunciation, and language-model pipelines with neural sequence models optimized directly from paired audio and text.
Connectionist temporal classification, attention-based encoder--decoders, and transducers are the dominant families.
RNN-Transducer (RNN-T) is particularly useful for streaming because it factors prediction into an acoustic encoder, a label-history prediction network, and a joint network that emits either a token or blank without requiring the complete utterance \cite{graves2012rnnt}.

The Conformer architecture combines self-attention, which captures long-range dependencies, with convolution, which captures local acoustic structure \cite{gulati2020conformer}.
FastConformer improves computational efficiency through aggressive subsampling and scalable attention variants \cite{rekesh2023fastconformer}.
Stateful and cache-aware Conformer inference reuses encoder states across chunks, avoiding full-prefix recomputation and permitting bounded future context \cite{noroozi2023stateful}.
These properties distinguish a genuinely streaming implementation from an application that repeatedly runs an offline model over overlapping windows.

Whisper demonstrates the robustness available from large-scale weak supervision, but its standard sequence-to-sequence formulation and fixed windows are optimized primarily for offline transcription rather than stateful token-by-token streaming \cite{radford2023whisper}.
This does not imply that Whisper cannot be used interactively; it means that latency, state, and recomputation trade-offs differ from an RNN-T designed for bounded-context inference.

\subsection{Multilingual and African ASR}

Multilingual pretraining can transfer acoustic and linguistic representations
to languages with limited labeled data. Ritchie et al. study multilingual
acoustic modeling and self-supervised transfer for African large-vocabulary
ASR \cite{dossou2022african}. Nahabwe et al. provide a unified comparison of
pretrained ASR model families across African languages and controlled data
scales \cite{african_asr_benchmark2025}.

The AfriVoices-KE resource expands this landscape with roughly 3,000 hours of
speech from five Kenyan languages and thousands of speakers
\cite{wanzare2026afrivoices}. Its scale is important, but large nominal
duration does not eliminate the need for split auditing and
transcription-quality analysis.

The present study differs from broad leaderboard work in scope.
It follows two systems from raw dataset preparation to deployment and retains a third unfinished system to expose the iterative failure modes.
The paper therefore complements benchmark studies: benchmarks compare model families on standardized data, while this work examines what happens between receiving a dataset and operating a language-specific streaming endpoint.

\subsection{Subword Modeling, Augmentation, and Prompt Conditioning}

SentencePiece learns subword units directly from text without language-specific tokenization rules \cite{kudo2018sentencepiece}.
Subword modeling reduces out-of-vocabulary failures and allows a multilingual model to share components across languages, although a tokenizer inherited from the base or bridge checkpoint may segment target-language forms inefficiently.
SpecAugment masks time and frequency regions in acoustic features and is a standard regularizer for ASR \cite{park2019specaugment}.
In this work, tokenizer and augmentation behavior are inherited from the restored checkpoint so that the first adaptation experiments change as few architectural variables as possible.

Nemotron 3.5 ASR represents the target-language choice as a 128-dimensional one-hot vector, broadcasts it across encoder time steps, concatenates it with the acoustic encoder output, and projects the fused representation before RNN-T decoding \cite{nvidia2026nemotron}. This input selects language-conditioned recognition behavior; it is not a translation instruction.
Our experiments retain the Kenyan Swahili prompt identifier \texttt{sw-KE} because the initialization checkpoint was trained with that carrier prompt.
The target-language evidence enters through paired target audio and text during full-model fine-tuning.

\section{System Architecture}

\subsection{Base Model}

The upstream model family is NVIDIA's \model{}, distributed as a NeMo model archive \cite{nvidia2026nemotron,nemo2019}. All target-language adaptations reported in this study initialize from Tony Kipkemboi's public Kenyan Swahili-adapted bridge checkpoint \cite{tonykip2026swahili}.

The restored bridge archive contains approximately 637 million trainable parameters, a 24-layer cache-aware FastConformer acoustic encoder with model dimension 1024, a 128-dimensional one-hot language identifier, and a SentencePiece inventory of 13,087 units. The recognizer uses an RNN-T objective and decoder. The NeMo archive packages the model weights, configuration, and tokenizer assets, supporting consistent restoration under a compatible NeMo software environment.

\begin{figure}[H]
\centering
\resizebox{0.98\textwidth}{!}{%
\begin{tikzpicture}[
  node distance=0.75cm and 0.85cm,
  box/.style={draw,rounded corners=2pt,minimum height=0.75cm,align=center,fill=CeloLight},
  small/.style={draw,rounded corners=2pt,minimum height=0.65cm,align=center,fill=white},
  arrow/.style={-{Latex[length=2.2mm]},thick}
]
\node[box] (audio) {Streaming\\audio chunks};
\node[box,right=of audio] (front) {Acoustic\\front end};
\node[box,right=of front,text width=2.7cm] (enc) {24-layer cache-aware\\FastConformer encoder\\($d=1024$)};
\node[small,above=of enc] (cache) {left-context and\\convolution caches};
\node[box,right=of enc] (fusion) {concatenation and\\prompt-fusion projection};
\node[box,right=of fusion] (joint) {RNN-T\\joint network};
\node[box,below=of joint] (pred) {RNN-T prediction\\network};
\node[small,below=of fusion] (prompt) {language-ID one-hot\\($K=128$; \texttt{sw-KE})};
\node[box,right=of joint] (tokens) {incremental\\subword tokens};
\draw[arrow] (audio) -- (front);
\draw[arrow] (front) -- (enc);
\draw[arrow] (enc) -- (fusion);
\draw[arrow] (fusion) -- (joint);
\draw[arrow] (pred) -- (joint);
\draw[arrow] (joint) -- (tokens);
\draw[arrow] (cache) -- (enc);
\draw[arrow] (enc) -- (cache);
\draw[arrow] (prompt) -- (fusion);
\draw[arrow] (tokens.south) |- (pred.east);
\end{tikzpicture}%
}
\caption{Conceptual architecture retained through adaptation and deployment. The encoder and decoder are stateful across chunks; the figure omits implementation-specific feature and subsampling details that remain serialized in the source checkpoint.}
\label{fig:architecture}
\end{figure}
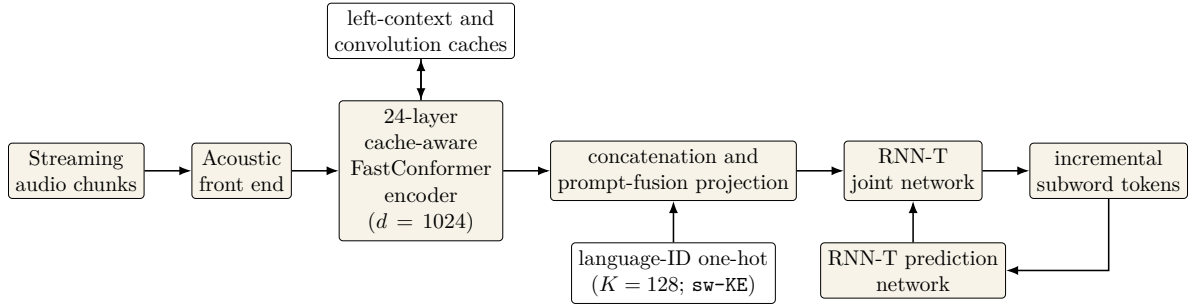
\FloatBarrier

\subsection{Cache-Aware Attention Context}

The model exposes an attention context pair written as \texttt{[L,R]}, where the project uses a large cached left context and bounded right context.
The deployed configuration is \texttt{[56,13]}. In the inspected model family, context units correspond to 80~ms encoder-frame increments.
The right-context value counts future units, while the model card's nominal chunk span includes the current unit. Consequently, \texttt{[56,13]} has approximately 1.04~s of future context and a 1.12~s nominal chunk span; \texttt{[56,3]} has approximately 240~ms of future context and a 320~ms nominal chunk span.
Larger right context generally improves recognition but delays commitment and increases the amount of future audio required before emitting stable output.

\begin{table}[H]
\caption{Attention-context configurations used during project experiments.}
\label{tab:contexts}
\centering
\begin{tabularx}{\textwidth}{lrrrY}
\toprule
\textbf{Context} & \textbf{Right units} &
\textbf{Future context} & \textbf{Nominal chunk span} & \textbf{Role} \\
\midrule
\texttt{[56,0]}  & 0  & 0 ms    & 80 ms  & minimum-future-context diagnostic \\
\texttt{[56,1]}  & 1  & 80 ms   & 160 ms & latency diagnostic \\
\texttt{[56,3]}  & 3  & 240 ms  & 320 ms & low-latency candidate \\
\texttt{[56,6]}  & 6  & 480 ms  & 560 ms & intermediate candidate \\
\texttt{[56,13]} & 13 & 1.04 s  & 1.12 s & selected quality configuration \\
\bottomrule
\end{tabularx}
\end{table}

The validation WER used for checkpoint ranking is produced by the standard NeMo trainer validation path over complete utterances; it is not the cache-aware true-streaming evaluator described in Section~\ref{sec:evaluation}.
True-streaming internal evaluation is run separately after checkpoint export.
For the historical Kalenjin v1-v run, the configured \texttt{validation.json} was additionally mixed-source, as disclosed in the data-split and Kalenjin-results sections.

The numerical context is only one part of latency. Perceived delay also depends on audio chunk size, microphone buffering, resampling, network transport, model cold start, GPU scheduling, decoder behavior, and endpointing.
For that reason, an evaluation runtime over a batch of files cannot be interpreted as per-utterance latency.

\subsection{RNN-T Objective}

Let $\mathbf{x}=(x_1,\ldots,x_T)$ be the acoustic feature sequence, $\mathbf{y}=(y_1,\ldots,y_U)$ the target subword sequence, and $\mathbf{q}_{\ell}$ the 128-dimensional one-hot language condition.
The prompt-fusion operation defined in Appendix~\ref{app:model-ledger} produces the language-conditioned acoustic representation $\widetilde{\mathbf{h}}^{\mathrm{enc}}_t$.
The RNN-T model sums over monotonic alignments $\mathcal{A}(\mathbf{x},\mathbf{y})$ containing output labels and blank symbols:
\begin{equation}
P(\mathbf{y}\mid\mathbf{x},\mathbf{q}_{\ell})=
\sum_{\mathbf{a}\in\mathcal{A}(\mathbf{x},\mathbf{y})}
\prod_{(t,u)\in\mathbf{a}}
P\!\left(a_{t,u}\mid
\widetilde{\mathbf{h}}^{\mathrm{enc}}_t,
\mathbf{h}^{\mathrm{pred}}_u\right).
\end{equation}
The base transducer term is
\begin{equation}
\mathcal{L}_{\mathrm{RNNT}}
=-\log P(\mathbf{y}\mid\mathbf{x},\mathbf{q}_{\ell}),
\end{equation}
and the reported implementation adds the FastEmit sequence-level emission regularizer \cite{yu2021fastemit}:
\begin{equation}
\mathcal{L}_{\mathrm{train}}
=\mathcal{L}_{\mathrm{RNNT}}+\lambda_{\mathrm{FE}}\mathcal{R}_{\mathrm{FE}},
\qquad \lambda_{\mathrm{FE}}=0.005.
\end{equation}
The exact regularizer is delegated to NeMo's RNN-T loss implementation. FastEmit encourages earlier label emission, which is useful for streaming but does not by itself guarantee good endpointing or partial-hypothesis stability.

\section{Why a Kenyan Swahili Bridge?}

Direct adaptation from a broad multilingual checkpoint is possible, but the project instead starts from a public Kenyan Swahili Nemotron 3.5 streaming checkpoint released by Tony Kipkemboi \cite{tonykip2026swahili}.
The rationale is practical rather than genealogical.
Swahili, Kikuyu, Dholuo, and Kalenjin do not all belong to one language subgroup, but Kenyan speech shares recording environments, names, loanwords, code-switch patterns, and regional acoustic conditions that may be absent from a globally averaged base model.
The bridge also demonstrates that the prompt-conditioned RNN-T can already be fine-tuned in the target NeMo environment.

The bridge is treated as an initialization, not as a lexical teacher.
The prompt remains \texttt{sw-KE}, while all trainable parameters are updated from target-language audio--transcript pairs.
Thus the output language is determined by the supervised target data and learned parameter updates; the prompt label does not cause target speech to be translated into Swahili.

This design creates an explicit experimental limitation.
The study does not yet include a controlled comparison against adaptation from the untouched NVIDIA base under identical seeds and compute.
Consequently, the bridge is a well-motivated engineering choice, not a proven causal improvement.
A future ablation should train base-to-target and bridge-to-target branches using the same manifests, optimizer state, step budget, and evaluator.

\section{Data Sources, Governance, and Split Semantics}

\subsection{Source Corpora}

The study uses African Next Voices (ANV) datasets distributed through Hugging Face for Kikuyu, Dholuo, and Kalenjin \cite{anv_kikuyu,anv_dholuo,anv_kalenjin,wanzare2026afrivoices}.
Each source provides audio, transcriptions, and metadata such as recording type, domain, and recorder identifier when available.
Both scripted and unscripted speech are present.

The manuscript reports effective post-cleaning data rather than only the row counts displayed on a dataset landing page.
Rows can disappear because audio is missing or undecodable, the transcript is empty, the duration exceeds the configured threshold, replacement characters make the text uncertain, or a later language-specific audit identifies non-speech artifacts.

\begin{table}[t]
\caption{Audited role-specific corpora in the final corpus definitions.
For Kalenjin, validation-development denotes the corrected disjoint split;
historical v1-v instead used the mixed-source selection manifest described
in Section~\ref{sec:results}. Hours are decoded audio duration after filtering.}
\label{tab:effective-data}
\centering
\small
\begin{tabular}{llrr}
\toprule
\textbf{Language} & \textbf{Split} & \textbf{Utterances} & \textbf{Hours} \\
\midrule
Kikuyu & train & 114,457 & 182.31 \\
        & validation & 13,667 & 19.64 \\
        & test & 6,795 & 10.62 \\
\midrule
Dholuo & train & 92,287 & 206.87 \\
       & validation & 11,152 & 25.19 \\
       & test & 5,480 & 12.16 \\
\midrule
Kalenjin clean-v3 & train & 64,007 & 151.28 \\
       & validation-dev & 8,953 & 18.92 \\
       & test & 2,411 & 5.33 \\
\bottomrule
\end{tabular}
\end{table}

For Kalenjin, we report the corrected, disjoint validation-development split containing 8,953 utterances. The historical v1-v run instead selected checkpoints using the mixed-source 11,364-row \texttt{validation.json};
see Section~\ref{sec:results}.

\subsection{Usage Rights and Community Governance}

The datasets contain human speech and should not be treated as anonymous numerical material. The raw ANV audio, speaker metadata, and other source-controlled materials are not redistributed by this work. Access, attribution, and permitted downstream use remain governed by the applicable dataset-provider terms. At the time of publication, the model repositories are maintained as private preservation mirrors rather than public releases.

Native-speaker review played a direct role in data decisions.
For Kikuyu, reviewers helped distinguish orthographic corruption from legitimate diacritics.
For Kalenjin, a reviewer confirmed that the lexical content of the reviewed examples was generally valid and that the exact forms \texttt{[cs]}, \texttt{(cs)}, \texttt{[pause]}, and \texttt{(pause)} were annotation wrappers rather than spoken words.
The later clean-v3 exclusion of rows containing the literal phrase \texttt{long pause} was a separate conservative, scope-defining project decision; it was not part of the speaker-confirmed marker set.
This review is not a substitute for broad community governance, but it prevented the pipeline from treating uncertain editorial assumptions as linguistic facts.

\subsection{Split Integrity}

Kikuyu and Dholuo use the official ANV train, validation, and test splits as the source of truth.
No validation or test row is included in gradient updates.
The Dholuo audit found zero recorder overlap across train, validation, and test in the metadata available to the project.

Kalenjin required an additional correction.
The raw validation package contained both \texttt{dev} and \texttt{dev\_test} source rows, while the raw test split corresponded to \texttt{dev\_test}.
Using the full raw validation split would therefore duplicate test material during model selection.
The corrected pipeline creates a disjoint \texttt{validation\_dev.json} from only the \texttt{dev} rows and reserves \texttt{dev\_test} for test.
Hash and provenance audits over the clean training validation and test manifests found zero overlap.

This correction describes the role-specific manifests produced by the current pipeline, but it was not used by the historical Kalenjin v1-v training command.
That run selected checkpoints with the mixed-source \texttt{validation.json}, which contained \texttt{dev\_test}-origin rows also represented in the clean-v3 test manifest.
Accordingly, its logged validation WER of 0.66772 is not a disjoint validation estimate, and the downstream clean-v3 test result is reported only as an adaptively consulted diagnostic.

This distinction is critical.
A split can be named ``validation'' and still violate the intended statistical role if its provenance contains a future test subset.
The project therefore records both the exported split name and the source-row split for every retained example.

\section{Data Preparation and Language-Specific Curation}

\subsection{Common Preparation Pipeline}

The common pipeline performs the following stages:
\begin{enumerate}
\item stream or load each source row while retaining source split, filename, recording type, domain, and recorder metadata;
\item decode the audio, reject missing or undecodable files, and measure duration from the decoded signal;
\item reject empty or unusable transcripts and rows beyond the maximum duration configured for the language and corpus generation;
\item apply a deterministic text normalizer and language-specific corrections;
\item write a NeMo JSON-lines manifest containing the audio path, duration, normalized text, and audit metadata;
\item persist per-reason skip counts, hours, type counts, domain counts, recorder counts, and normalization-change counts;
\item package audio into tarred shards for efficient training while preserving manifest-to-shard correspondence;
\item run split-overlap and file-existence audits before dispatching a GPU training job.
\end{enumerate}

\begin{table}[H]
\caption{Maximum accepted duration in the final reported corpus generation for each language.}
\label{tab:duration-limits}
\centering
\begin{tabular}{lr}
\toprule
\textbf{Language} & \textbf{Maximum accepted duration} \\
\midrule
Kikuyu & 30.0 s \\
Dholuo & 39.99 s \\
Kalenjin & 39.99 s \\
\bottomrule
\end{tabular}
\end{table}

The executable normalizer is the canonical specification.
The implementation records the normalization rules and corpus ledgers. A future public release should additionally include unit tests with before-and-after strings because Unicode punctuation and visually similar letters are easy to misdescribe in prose.
Rows with uncertain text are dropped rather than rewritten when the project lacks enough linguistic evidence to assert the intended form.

\subsection{Dholuo Preparation Accounting}

The raw Dholuo training pass observed 117,564 rows and retained 92,287.
The retained audio totals 206.8679 hours and includes 85,394 scripted and 6,893 unscripted examples.
The preparation report records 1,913 rows with replacement characters, 41 decode failures, 11,159 over-duration rows, and 12,164 empty transcriptions.
Some rows can trigger more than one diagnostic during exploration; the production manifest writer assigns a final exclusion reason deterministically.

The raw Dholuo validation pass observed 13,853 rows and retained 11,152 (25.1899 hours).
It excluded 221 replacement-character rows, 1,138 over-duration rows, three decode failures, and 1,339 empty transcriptions.
The retained split contains 10,271 scripted and 881 unscripted examples.
The test set retains 5,480 examples totaling 12.16 hours.

\begin{table}[H]
\caption{Dholuo preparation ledger for the train and validation splits.}
\label{tab:dholuo-prep}
\centering
\small
\begin{tabular}{lrr}
\toprule
\textbf{Quantity} & \textbf{Train} & \textbf{Validation} \\
\midrule
Rows observed & 117,564 & 13,853 \\
Rows retained & 92,287 & 11,152 \\
Retained hours & 206.8679 & 25.1899 \\
Replacement-character exclusions & 1,913 & 221 \\
Decode failures & 41 & 3 \\
Over-duration exclusions & 11,159 & 1,138 \\
Empty-transcript exclusions & 12,164 & 1,339 \\
Scripted retained & 85,394 & 10,271 \\
Unscripted retained & 6,893 & 881 \\
\bottomrule
\end{tabular}
\end{table}

\subsection{Kikuyu Orthography and Unicode Decisions}

Kikuyu curation required special attention to the contrastive characters \~{\i} and \~{u} as represented in the project's Unicode text.
Early audits found visually related but unintended forms such as \texttt{å}, \texttt{ä}, \texttt{ï}, and \texttt{\={u}} in places where the intended Kikuyu character could be established from project evidence.
The reviewed correction path included \texttt{å}$\rightarrow$\texttt{\~{u}}, \texttt{ä}$\rightarrow$\texttt{\~{\i}}, followed by \texttt{ï}$\rightarrow$\texttt{\~{\i}} and \texttt{\={u}}$\rightarrow$\texttt{\~{u}} for confirmed cases.
Accepted text is serialized in Unicode Normalization Form C (NFC) so that canonically equivalent character sequences do not create distinct labels.

The important methodological choice is what happened when the mapping was not certain.
The pipeline did not apply arbitrary transliteration merely to retain more hours.
Rows containing unresolved suspicious characters were excluded in the final curated generation.
This reduced nominal data volume but prevented the model from learning a mixture of valid orthography and encoding noise.

The final filtered-v4 Kikuyu corpus contains 114,457 training utterances
(182.31 hours), 13,667 validation utterances (19.64 hours), and 6,795
internal-test utterances (10.62 hours). The validation and test splits were
excluded from gradient updates but were consulted adaptively during the
iterative research process.

\subsection{Kalenjin: From Clean-v1 to Clean-v3}

The first Kalenjin preparation pass observed 82,378 training rows and retained 68,012 (158.9356 hours).
It excluded 450 rows with replacement characters, 392 decode failures, 12,621 over-duration rows, one transcript empty after normalization, and 902 empty transcriptions.
The retained data included 58,927 scripted and 9,085 unscripted examples.

The first validation pass observed 13,845 rows and retained 12,084 (25.5004 hours), while the first test pass observed 3,315 and retained 2,557 (5.6431 hours).
These manifests were useful for early learning experiments but retained annotation markers that punctuation normalization converted into apparent words.

Clean-v2 explicitly removes the confirmed marker forms \texttt{[cs]}, \texttt{(cs)}, \texttt{[pause]}, and \texttt{(pause)}.
An evaluation of the existing v1-iii checkpoint against clean-v2 separated metric changes caused by reference cleanup from those caused by learned parameters.
However, further audit found the literal phrase \texttt{long pause}, digit-bearing references such as \texttt{20082018} and \texttt{we5107}, and very short normalized targets associated with over-generation.

Clean-v3 derives from clean-v2 and applies three row-level filters: it drops the \emph{entire row} when the normalized reference contains the literal phrase \texttt{long pause}, when it contains any digit, or when it contains fewer than three normalized tokens.
The digit rule intentionally excludes all digit-bearing references in this experimental generation, including forms that may be linguistically recoverable; it should not be interpreted as evidence that every number or identifier is invalid speech.
The pipeline does not merely delete a suspicious substring and retain the remaining words, because doing so could create audio--text misalignment.
The resulting clean-v3 corpus contains 64,007 training utterances (151.28 hours), 8,953 disjoint validation-development utterances (18.92 hours), and 2,411 test utterances (5.33 hours).

\begin{table}[H]
\caption{Kalenjin corpus generations and their purpose.}
\label{tab:kalenjin-cleaning}
\centering
\small
\begin{tabularx}{\textwidth}{L{1.5cm}rrrY}
\toprule
\textbf{Version} & \textbf{Train} & \textbf{Val.} & \textbf{Test} & \textbf{Primary change} \\
\midrule
clean-v1 & 68,012 & 12,084 & 2,557 & Common decode, duration, replacement-character, and empty-text filters. Raw validation still required provenance separation. \\
clean-v2 & --- & --- & 2,557 & Removed confirmed bracketed/parenthesized \texttt{cs} and \texttt{pause} markers; created disjoint development validation. \\
clean-v3 & 64,007 & 8,953 & 2,411 & Dropped rows containing literal \texttt{long pause}, any digit, or fewer than three normalized tokens. \\
\bottomrule
\end{tabularx}
\end{table}

\subsection{Why Deletion Was Sometimes Preferable to Repair}

Deleting an entire row sacrifices data, but partial text repair can be more damaging when the suspicious token corresponds to audible material of unknown extent.
For example, deleting \texttt{long pause} from the text is safe only if it is a non-speech annotation and not a phrase read aloud.
Deleting \texttt{20082018} is unsafe if the speaker actually says a date whose words should be represented.
Without row-level listening and language-aware transcription, retaining the rest of the text can create a target that is systematically shorter than the audio.

The project uses three decision levels:
\begin{enumerate}
\item deterministic normalization for cases whose intended character or marker interpretation is verified;
\item row-level native-speaker review for ambiguous but potentially valuable cases;
\item conservative row exclusion when the intended supervision cannot be reconstructed reliably.
\end{enumerate}

This policy is intentionally reversible.
The raw data and exclusion report are preserved so that a later transcription campaign can recover excluded audio with corrected labels.

\section{Implementation and Experimental Infrastructure}
\label{sec:implementation}

\subsection{Software Stack}

The training and inference system is implemented in Python using PyTorch through NVIDIA NeMo's ASR abstractions \cite{nemo2019}.
NeMo supplies the model restoration, tarred-audio data loaders, prompt-aware RNN-T model class, mixed-precision execution, distributed-training integration, checkpoint callbacks, and cache-aware streaming inference utilities.
PyTorch remains the numerical and automatic-differentiation substrate: all trainable tensors, optimizer state, gradient computation, and GPU kernels ultimately execute through PyTorch.

The remote training jobs run in project-defined, version-controlled container image configurations on Modal GPU workers.
Weights, manifests, tarred audio shards, logs, evaluation reports, and exported \texttt{.nemo} archives are stored in persistent Modal volumes rather than ephemeral container filesystems.
Weights \& Biases records optimization metrics and validation trajectories.
The deployment layer uses language-specific FastAPI WebSocket services
for the isolated true-streaming backends. The browser client obtains language-specific
endpoint configuration through a protected application route and then
opens the configured stream; it never receives storage credentials or
checkpoint paths.

\subsection{Persistent Data Layout}

Each language has an isolated persistent volume with a common logical layout:
\begin{lstlisting}[language={},caption={Logical persistent-volume layout used by each language project.}]
/
  anv_<language>/
    <corpus_generation>/
      manifests/{train,validation,test}.json
      audio/{train,validation,test}/...
      shards/...
      manifest_summary.json
  checkpoints/<run_name>/<run_name>/<timestamp>/checkpoints/...
  eval/<evaluation_name>/...
  reports/...
  logs/<run_name>/...
  models/...
\end{lstlisting}
The container mounts this volume at \path{/data}; consequently, a volume path such as \path{/checkpoints/...} appears inside a running worker as \path{/data/checkpoints/...}.
This distinction became operationally important when verifying paths with the Modal CLI: the CLI lists paths relative to the volume root, while training code sees the mount point.

\subsection{Why Tarred Audio Was Used}

Reading tens of thousands of independent audio objects from a remote persistent filesystem creates metadata and open-call overhead.
The pipeline therefore supports tarred WebDataset-style audio shards, with manifests carrying shard assignments and NeMo reading sequentially from archives.
This improved input throughput, but introduced a strict integrity requirement: every manifest entry, shard identifier, archive name, and row count had to agree.
Early failures caused by stale shard metadata or missing \texttt{shard\_id} fields were treated as data-build failures rather than training failures.
The production rule became: generate manifests and shards in one versioned operation, validate all members, and never mix a manifest from one generation with archives from another.

\subsection{Compute Selection}

Short probes were used to validate compatibility before committing expensive accelerator time.
Full experiments used NVIDIA H100 or H200 workers depending on availability and memory requirements.
The H200 was chosen for the principal full-data Dholuo and Kalenjin continuation runs because its larger memory headroom accommodated the 0.6B model, duration-based batches, validation decoding, and checkpoint serialization without changing the optimization objective.
The accelerator choice is therefore an infrastructure variable, not a claimed modeling contribution.

\subsection{Run Dispatch and Lifecycle}

Long jobs were dispatched with Modal's detached execution mode so that a laptop sleep, terminal closure, or transient network failure would not cancel the remote input.
This was not merely ergonomic.
During development, a foreground \texttt{remote()} call lost its gRPC connection and Modal propagated a cancellation signal to the active training input.
The corrected production invocation used \texttt{modal run -{}-detach} and an entrypoint that retained the spawned function call independently of the local client lifecycle.

Some successful jobs ended with a Modal message stating that the runner had been shutting down longer than its grace period.
Such a message was not interpreted in isolation.
A run was considered successfully trained only when the logs showed that the target step was reached, the final checkpoint and \texttt{.nemo} archive were saved, the persistent volume was committed, and the W\&B run closed.
The post-completion runner warning was recorded as an infrastructure incident, not silently discarded.

\section{Fine-Tuning Protocol}
\label{sec:training}

\subsection{Full-Parameter Adaptation}

All reported language models use full-parameter fine-tuning rather than adapters or a frozen acoustic encoder.
Let \(\theta\) denote all acoustic encoder, prompt-fusion projection, prediction-network, and joint-network parameters.
Given an audio sequence \(\mathbf{x}\), one-hot language identifier \(\mathbf{q}_{\ell}\), and target subword sequence \(\mathbf{y}\), training minimizes the RNN-T negative log likelihood
\begin{equation}
\mathcal{L}_{\mathrm{RNNT}}(\theta)
=-\log \sum_{\pi \in \mathcal{B}^{-1}(\mathbf{y})}
P_{\theta}(\pi\mid\mathbf{x},\mathbf{q}_{\ell}),
\label{eq:rnnt-loss}
\end{equation}
where \(\pi\) ranges over blank-augmented alignments that collapse to \(\mathbf{y}\).
The implementation uses \texttt{warprnnt\_numba} and a FastEmit coefficient of 0.005 to encourage timely token emission without replacing the transducer objective \cite{yu2021fastemit}.

\subsection{Core and Stage-Specific Hyperparameters}

The continuation lineage reused a recurring low-learning-rate
configuration, but warm-up, checkpoint retention, and some controlled
challenger settings were not invariant across every run.
Table~\ref{tab:shared-hparams} distinguishes the core configuration of
the selected fixed-rate late-stage path from stage-specific settings
explicitly preserved in the project records. Exact values are not
inferred for stages whose launch configurations were not explicitly
recorded in the inspected artifacts.

\begin{table}[H]
\caption{Core and stage-specific configuration documented for selected
late-stage continuation runs.}
\label{tab:shared-hparams}
\centering
\small
\begin{tabularx}{\textwidth}{L{4.2cm}Y}
\toprule
\textbf{Component} & \textbf{Configuration} \\
\midrule
Model update & Full-parameter fine-tuning \\
Optimizer & AdamW \\
Learning rate & \(5\times10^{-7}\) for the selected fixed-rate continuation branches \\
Scheduler minimum & \(5\times10^{-7}\) for those fixed-rate branches \\
Weight decay & 0.001 \\
Training batch construction & Duration-based, normally 200 seconds of aggregate audio in the documented late-stage runs \\
Validation batch construction & Duration-based, normally 120 seconds of aggregate audio \\
Validation interval & Every 1,000 optimizer steps over the complete model-selection validation manifest \\
Warm-up and checkpoint retention &
Kikuyu \texttt{v9\_i} and the v10 branches used 500 warm-up steps and retained the best five checkpoints; Kikuyu v11 used 500 warm-up steps and retained the best ten; Dholuo v6--v7 and Kalenjin v1-iv--v1-v used 100 warm-up steps and retained the best ten \\
Loss & RNN-T via \texttt{warprnnt\_numba} \\
FastEmit coefficient & 0.005 \\
Prompt & \texttt{sw-KE} \\
Streaming attention context & \texttt{[56,13]} for the reported evaluations \\
Decoding & Greedy cache-aware streaming RNN-T \\
Precision and accelerator & NeMo/PyTorch mixed-precision GPU execution; H100/H200 according to run \\
\bottomrule
\end{tabularx}
\end{table}

\subsection{Staged Continuation}

The project uses a checkpoint lineage: a selected model from one stage initializes the next stage.
This is useful operationally because each stage can be evaluated before additional compute is spent.
It also permits corpus corrections between stages, as in Kalenjin clean-v2 and clean-v3.

Staged continuation must not be described as mathematically identical to a single uninterrupted run of the combined length.
Restoring a \texttt{.nemo} model archive restores model parameters and configuration, but the continuation scripts initialize a new optimizer and scheduler unless a complete trainer checkpoint is resumed.
The data order, warm-up, random state, and optimizer moments can therefore differ.
In this paper, a statement such as ``v7 continues from v6'' means parameter initialization from the selected v6 model, not exact continuation of v6's optimizer trajectory.

\subsection{Checkpoint Selection and Export}

Each full validation produces a validation-WER measurement.
The checkpoint callback retains a run-specific top-\(k\) subset of
checkpoint candidates. The selection rule is
\begin{equation}
k^*=\arg\min_{k\in\mathcal{K}} \wer_{\mathrm{validation}}(k),
\label{eq:checkpoint-select}
\end{equation}
where \(\mathcal{K}\) is the set of checkpoints retained by the callback of that run
.
The target promotion procedure restores the selected checkpoint and exports it to a self-contained NeMo \texttt{.nemo} archive.
That archive bundles the model configuration, tokenizer assets, and weights required by NeMo restoration.

The only multi-checkpoint averaging result reported here is from the Kikuyu v4 filtered experiment.
The selected individual checkpoint obtained 53.3505\% WER and 10.7256\% NS-CER, whereas the top-three checkpoint average obtained 53.3887\% WER and 10.7683\% NS-CER under the same filtered full-test protocol.
The top-three average therefore did not improve upon the selected individual checkpoint. We did not conduct additional ablations to determine the cause of this result.
For v13, we exported the validation-selected best individual checkpoint, denoted \emph{best1}. The archived v13 experiment record contains no checkpoint-averaging evaluation, so we report only the individual checkpoint result.

Dholuo v7 is a documented historical exception to this export procedure.
The run reached a best validation WER of 0.33944, but the held-out evaluation and preserved deployment package used the final v7 \texttt{.nemo} export, whose final logged validation WER was 0.33949.
The best-validation trainer checkpoint was preserved separately and was not the artifact used to obtain the reported Dholuo internal evaluation metrics.

\section{Evaluation Protocol}
\label{sec:evaluation}

\subsection{Metrics}

For a reference containing \(N\) words, word error rate is
\begin{equation}
\wer=\frac{S+D+I}{N},
\label{eq:wer}
\end{equation}
where \(S\), \(D\), and \(I\) are the minimum-edit substitution, deletion, and insertion counts.
Character error rate is computed analogously over Unicode characters:
\begin{equation}
\cer=\frac{S_c+D_c+I_c}{N_c}.
\label{eq:cer}
\end{equation}
Because inconsistent word-boundary decisions can dominate strict WER in agglutinative or under-standardized text, the project also reports no-space CER:
\begin{equation}
\nscer=\cer\bigl(\rho(r),\rho(h)\bigr),
\qquad \rho(s)=\text{remove-spaces}(s),
\label{eq:nscer}
\end{equation}
for reference \(r\) and hypothesis \(h\).
No-space CER is complementary; it does not replace WER or standard CER.

\subsection{True-Streaming Evaluation}

This post-training evaluator is distinct from the full-utterance trainer validation used to rank checkpoints.
The internal true-streaming evaluator uses the cache-aware streaming path rather than transcribing complete files through an offline convenience API.
Audio is segmented into successive chunks according to the model's streaming configuration.
For each active utterance, the evaluator carries forward the encoder cache, cache lengths, and RNN-T decoding state.
Only the current chunk and permitted right context are available when producing the next partial result.
The final hypothesis is scored after the utterance is flushed.

Unless otherwise stated, reported tests use prompt \texttt{sw-KE}, attention context \texttt{[56,13]}, greedy decoding, and the language's versioned internal evaluation manifest.
This alignment between evaluation and deployment is important: an offline WER can conceal regressions caused by chunking, cache management, or emission timing.

\subsection{Latency and Runtime Interpretation}

The evaluator records total wall-clock runtime for the complete test set.
This is useful for detecting gross throughput regressions, but it is not an end-user latency benchmark.
The runs include loading, batching, data access, Python overhead, and aggregate GPU execution.
A production latency study should separately report cold-start time, real-time factor, first-token latency, partial-update cadence, finalization delay, concurrency, and hardware.
Accordingly, this paper calls the measured quantity \emph{evaluation runtime}, not streaming latency.

\subsection{Held-Out Status and Repeated Use}

The Kikuyu and Dholuo test manifests were not used for gradient updates.
However, repeated model decisions were informed by full-test results across successive continuation stages.
The sets are therefore best described as \emph{internal held-out evaluation sets}, not untouched one-shot benchmarks.
This distinction limits the strength of external generalization claims.
A future public release should freeze an additional speaker-disjoint test set or use an independent benchmark whose references and normalization are never consulted during development.

\subsection{Comparability to External Leaderboards}

Numbers from public leaderboards are meaningful only when audio, reference normalization, segmentation, language variety, decoding mode, and metric implementation match.
The internal C-elo results are not claimed to be directly paired with PazaBench \cite{microsoft2026paza} or other public scores.
They can motivate a controlled external evaluation, but they do not by themselves establish state of the art.
This paper therefore reports exact internal protocols and avoids substituting visual leaderboard comparison for a reproduced benchmark.

\section{Experimental Results}
\label{sec:results}

\subsection{Kikuyu Learning Trajectory}

The Kikuyu lineage combined increasingly strict orthographic curation with staged continuation.
Table~\ref{tab:kikuyu-lineage} reports the internal true-streaming results for the principal retained stages.
Later stages use the same evaluation configuration and demonstrate a consistent reduction in both WER and no-space CER.

\begin{table}[H]
\caption{Kikuyu internal true-streaming results. Lower is better.}
\label{tab:kikuyu-lineage}
\centering
\small
\begin{tabular}{lrrl}
\toprule
\textbf{Checkpoint} & \textbf{WER (\%)} & \textbf{No-space CER (\%)} & \textbf{Role} \\
\midrule
v4 filtered & 53.3505 & 10.7256 & Orthography-filtered reference point \\
v6 & 51.2816 & 10.1190 & Continued adaptation \\
v7-i & 50.8278 & 9.9395 & Continued adaptation \\
v8-ii & 49.3441 & 9.5613 & Continued adaptation \\
v9 & 48.5926 & 9.3104 & Continued adaptation \\
v10 & 46.8393 & 8.7815 & Continued adaptation \\
v11 & 45.6926 & 8.4489 & Continued adaptation \\
v12 best1 & 44.6985 & 8.1824 & Prior production candidate \\
v13 best1 & \textbf{42.9677} & \textbf{7.7947} & Selected Kikuyu champion \\
\bottomrule
\end{tabular}
\end{table}

The v13 run continued from v12 for 70,000 optimizer steps.
Its best validation WER was 0.41163 near epoch 65; the final validation at step 70,000 was 0.41244.
This small late-stage regression is why the best checkpoint, not the final checkpoint, was exported.
Relative to v12, v13 reduced internal-evaluation WER by 1.7308 absolute points (3.87\% relative) and no-space CER by 0.3877 absolute points (4.74\% relative).
Relative to the v4 filtered reference, the cumulative absolute reductions were 10.3828 WER points and 2.9309 no-space CER points.

\subsubsection{Streaming context ablation}

An internal ablation compared \texttt{[56,13]} with \texttt{[56,3]} while holding the checkpoint and evaluation pipeline fixed.
The larger right-context setting improved WER by approximately 0.90 absolute points and no-space CER by approximately 0.40 points.
The deployed default was therefore fixed at \texttt{[56,13]}.
This is an accuracy-oriented operating point and should not be interpreted as the only valid latency setting.

\subsection{Dholuo Learning Trajectory}

Dholuo exhibits a particularly clear continuation curve.
Validation WER fell from 0.49586 in the first full-data stage to 0.33944 in v7.
Table~\ref{tab:dholuo-lineage} separates the validation-selection trajectory from held-out evaluations available for the final stages.

The historical Dholuo preparation pipeline did not remove the literal token
\texttt{cs}. Normalization removed surrounding brackets or parentheses but
did not remove the resulting token. Run logs directly confirm its presence
in validation references, and the same frozen normalization function was
used when constructing the train, validation, and test manifests.

Comparisons among checkpoints are interpretable only within the same
evaluation split and under the same label policy, decoder, attention
context, and scoring implementation. Validation metrics are used for model
selection, whereas test metrics summarize a separately defined evaluation
population; the two are not compared numerically with one another. Results
under this historical Dholuo policy should not be compared directly with
scores produced after removing \texttt{cs} or applying a different
normalization policy.

\begin{table}[H]
\caption{Dholuo full-data continuation lineage. Best validation WER reports
checkpoint-selection results. The v7 test metrics were measured from the
final exported \texttt{.nemo}, whose final logged validation WER was
0.33949; the separately preserved best-validation checkpoint reached
0.33944. An em dash denotes an unreported metric.}
\label{tab:dholuo-lineage}
\centering
\small
\begin{tabular}{lrrrrr}
\toprule
\textbf{Stage} & \textbf{Steps} & \textbf{Best val. WER} & \textbf{Test WER (\%)} & \textbf{CER (\%)} & \textbf{NS-CER (\%)} \\
\midrule
v3-i & 20,000 & 0.49586 & --- & --- & --- \\
v4-i & 70,000 & 0.38395 & --- & --- & --- \\
v5-i & 40,000 & 0.36139 & --- & --- & --- \\
v6-i & 20,000 & 0.35174 & 35.3502 & 9.8512 & 8.3031 \\
v7-i & 40,000 & \textbf{0.33944} & \textbf{33.9791} & \textbf{9.5855} & \textbf{8.1340} \\
\bottomrule
\end{tabular}
\end{table}

The v7 final export's full internal true-streaming evaluation produced
33.9791\% WER, 9.5855\% standard CER, and 8.1340\% no-space CER on
5,480 utterances.
The complete evaluation took 222.45 seconds on its evaluation worker.
Under the same Dholuo test manifest and scoring protocol, v7 reduced WER
relative to v6 by 1.3711 absolute points (3.88\% relative), standard CER by 0.2657 points, and no-space CER by 0.1691 points.
The smaller CER change relative to WER suggests that some of the remaining errors concern segmentation and word boundaries rather than entirely incorrect character sequences.

The v7 validation curve continued to decline through most of the run but had nearly plateaued by 40,000 steps.
This, together with the diminishing test improvement, motivated treating v7 as the final Dholuo champion rather than automatically launching another long continuation.

\subsection{Kalenjin Work in Progress}

Kalenjin remains an active research track rather than a production claim.
The lineage demonstrates that the model is learning, while also showing a large residual error burden and the importance of corpus revision.

\begin{table}[H]
\caption{Kalenjin adaptively consulted diagnostic progression. Corpus generations differ, and v1-v used mixed-source checkpoint selection, so adjacent rows should be interpreted with their stated reference set and limitations.}
\label{tab:kalenjin-lineage}
\centering
\small
\begin{tabularx}{\textwidth}{L{1.6cm}L{2.2cm}rrrY}
\toprule
\textbf{Stage} & \textbf{Corpus} & \textbf{WER (\%)} & \textbf{CER (\%)} & \textbf{NS-CER (\%)} & \textbf{Interpretation} \\
\midrule
v1-i & clean-v1 & 85.78 & --- & --- & First full-data bridge adaptation. \\
v1-ii & clean-v1 & 78.0242 & 24.4985 & 22.8378 & Continued learning under original references. \\
v1-iii & clean-v1 & 73.2007 & 21.9547 & 20.6846 & Best pre-clean-v2 evaluation. \\
v1-iv & clean-v2 & 70.7831 & 20.4366 & 19.4145 & Continued model plus cleaned marker references. Best validation WER 0.68255. \\
v1-v & clean-v3 & \textbf{68.74} & --- & --- & Clean-v3 true-stream diagnostic; mixed-source checkpoint-selection WER 0.66772. \\
\bottomrule
\end{tabularx}
\end{table}

The v1-v run initialized from v1-iv and trained for 70,000 steps on clean-v3.
Its best and final logged selection WER were both 0.66772, but the historical command used \texttt{validation.json} rather than the disjoint \texttt{validation\_dev.json}.
That mixed-source manifest contained \texttt{dev\_test}-origin rows also represented in the clean-v3 test set; therefore 0.66772 is not a disjoint validation estimate.
The true-streaming clean-v3 diagnostic produced 68.74\% WER on 2,411 utterances and took 92.16 seconds.
This filtered set excludes rows containing \texttt{long pause}, any digit, or fewer than three normalized tokens.
Because the set was repeatedly consulted during development and checkpoint selection had test-origin exposure, 68.74\% is an adaptive diagnostic result rather than an independent generalization estimate.
Because the evaluator output for this stage did not archive standard and no-space CER, those cells remain unreported rather than reconstructed from incomplete logs.

The Kalenjin result is not yet comparable in maturity to Kikuyu or Dholuo.
It documents a repeatable internal diagnostic after marker and artifact cleanup, and it identifies where native-speaker-guided error correction should focus.

\subsection{Cross-Language Snapshot}

\begin{table}[H]
\caption{Current project snapshot under the internal true-streaming
protocols. Kikuyu and Dholuo use adaptively consulted internal evaluation
sets; Dholuo uses its frozen historical normalization policy. Kalenjin is
a filtered adaptive diagnostic with mixed-source checkpoint-selection
exposure. These metrics are not substitutes for evaluation on a common
external benchmark.}
\label{tab:cross-language}
\centering
\small
\begin{tabularx}{\textwidth}{@{}lrrrrY@{}}
\toprule
\textbf{Language} & \textbf{Train h} & \textbf{Test rows} & \textbf{WER (\%)} & \textbf{NS-CER (\%)} & \textbf{Status} \\
\midrule
Kikuyu & 182.31 & 6,795 & 42.9677 & 7.7947 & Champion preserved/deployed \\
Dholuo & 206.87 & 5,480 & 33.9791 & 8.1340 & Champion preserved/deployed \\
Kalenjin & 151.28 & 2,411 & 68.74 & --- & Adaptive diagnostic; research in progress \\
\bottomrule
\end{tabularx}
\end{table}

The apparent ordering across languages must not be read as a controlled language-difficulty comparison.
The corpora differ in speakers, domains, scripted/unscripted balance, orthographic variation, utterance length, and reference quality.
The table summarizes project state; it does not establish that one language is intrinsically easier than another.

\section{Error Analysis}
\label{sec:error-analysis}

\subsection{Why Aggregate WER Is Insufficient}

An ASR system can improve aggregate WER while remaining unreliable for a particular speaker group, recording condition, domain, or utterance type.
The project therefore treats WER as the beginning of diagnosis rather than the end.
For each evaluated utterance, the analysis table can retain reference, hypothesis, word-edit counts, character-edit counts, duration, scripted/unscripted type, domain, recorder identifier, and source split.
This enables both quantitative slicing and native-speaker review.

The most useful distinction is between an \emph{acoustic error}, where the hypothesis is not phonetically faithful; an \emph{orthographic error}, where a plausible sound sequence is written differently; a \emph{boundary error}, where neighboring morphemes or words are joined or split; and an \emph{annotation mismatch}, where the reference contains metadata or text that was not spoken.
These categories have different remedies.
More optimization cannot reliably fix a corrupted reference, while text normalization alone cannot fix poor acoustic coverage.

\subsection{Kikuyu and Dholuo Boundary Sensitivity}

The gap between strict WER and no-space CER is informative.
Kikuyu v13 has 42.9677\% WER but 7.7947\% no-space CER; Dholuo v7 has 33.9791\% WER and 8.1340\% no-space CER.
This does not mean that the models make only 8\% meaningful errors, because removing spaces intentionally ignores word boundaries.
It does show that many hypotheses preserve a substantial portion of the character sequence even when strict tokenization yields substitutions, insertions, or deletions.

For example, a reference may contain a compound as one token while the model emits two acoustically close tokens.
Strict WER can score this as multiple edits; no-space CER reduces the penalty if the underlying character stream is similar.
For downstream search or subtitle display, the distinction matters: a near-phonetic string may be understandable to a speaker while still being unsuitable for a lexically normalized transcript.

\subsection{Kalenjin Stratified Findings}

The detailed v1-iii audit exposed several systematic patterns before clean-v3 was constructed:
\begin{itemize}
\item scripted utterances had approximately 70.02\% WER, while unscripted utterances reached approximately 83.61\%;
\item utterances longer than 41 reference words reached approximately 84.54\% WER;
\item audio in the 20--40 second duration bucket reached approximately 83.27\% WER;
\item only 68 of 2,557 test rows contained the literal \texttt{cs} marker, and removing those rows changed aggregate WER from approximately 73.20\% to 73.03\%;
\item frequent errors involved word joining and splitting, for example a reference form such as \texttt{inyaaksei} versus a hypothesis sequence resembling \texttt{inya aksei}.
\end{itemize}

The small effect of excluding \texttt{cs} rows is an important negative result.
Marker cleanup was necessary for label validity, but it could not explain the majority of Kalenjin errors.
The large scripted/unscripted and long-utterance gaps point instead to acoustic and linguistic coverage, transcript consistency, and sequence-length robustness.

\subsection{Short-Utterance Over-Generation}

Very short references can produce high insertion rates when the model emits a longer, plausible sequence.
Several causes are possible: incorrect end-of-utterance boundaries, residual non-speech in the audio, references that omit spoken material, prompt or language-model priors that dominate weak acoustics, and genuine model over-generation.
The clean-v3 decision to exclude normalized targets shorter than three tokens is therefore an experimental control, not a universal recommendation.
It reduces a concentrated source of instability while an explicit endpointing and label audit is performed.

A production remedy should not simply discard all short speech.
The stronger next study is a short-utterance challenge set with manual listening, verified time boundaries, speech-activity metadata, insertion/deletion breakdowns, and inference-side endpoint tuning.

\subsection{Code-Switching and Borrowed Terms}

The corpora contain English and Swahili names, products, institutions, and technical terms embedded in the target language.
Examples include service names, crop names, healthcare terminology, and public institutions.
In the Kalenjin clean-v2 and clean-v3 label generations, the speaker-confirmed annotation wrappers \texttt{[cs]}, \texttt{(cs)}, \texttt{[pause]}, and \texttt{(pause)} were removed, while genuinely spoken code-switched words were retained.
Clean-v3 additionally rejected complete rows containing the literal annotation phrase \texttt{long pause}.
The historical Dholuo series is an explicitly disclosed exception: its
normalizer retained the unbracketed token \texttt{cs} wherever it occurred,
and run logs directly confirm it in validation targets. The same frozen
normalizer was used to construct all three data splits.

Errors around borrowed terms should be analyzed separately from errors on core target-language words.
The inherited SentencePiece vocabulary can compose many previously unseen strings from subword pieces, subject to its fixed inventory, normalization, and unknown-token behavior; frequency and orthographic priors still matter.
A future tokenizer study should compare the inherited vocabulary with a jointly trained tokenizer while keeping the acoustic model and data fixed.

\subsection{Representative Qualitative Patterns}

In Table~\ref{tab:qualitative-patterns}, the reference excerpt is taken from the dataset transcript associated with the audio, whereas the hypothesis excerpt is the corresponding text predicted by the ASR model. The examples are diagnostic snippets rather than linguistically adjudicated judgments hence fluent speakers should determine which differences are acceptable variants.

\begin{table}[H]
\caption{Representative qualitative patterns from true-streaming logs.}
\label{tab:qualitative-patterns}
\centering
\scriptsize
\begin{tabularx}{0.98\textwidth}{L{1.6cm}Y Y L{2.5cm}}
\toprule
\textbf{Language} & \textbf{Reference excerpt} & \textbf{Hypothesis excerpt} & \textbf{Likely pattern} \\
\midrule
Dholuo & \texttt{tijni nigi ber mogundho ahinya} & \texttt{tijni nigi ber mogwundho ahinya} & Local character substitution \\
Dholuo & \texttt{gitero e chiro ka giuso} & \texttt{gitero e chiro ka gi uso} & Word-boundary difference \\
Dholuo & \texttt{cashew nuts} & forms resembling \texttt{cashonarts} & Borrowed-term acoustic/tokenizer error \\
Kalenjin & \texttt{huduma center} & \texttt{huduma synda} & Borrowed proper/service term \\
Kalenjin & \texttt{immunotherapy} & \texttt{immuno therapy} & Boundary difference in technical term \\
Kalenjin & \texttt{kibedi keleweni kodoik kyok} & \texttt{kendi kelewini kandoityo} & Multiple lexical/acoustic substitutions \\
Kikuyu & one orthographic token & acoustically close split or joined forms & Boundary and orthography interaction \\
\bottomrule
\end{tabularx}
\end{table}

\subsection{Error-Analysis Priorities}

Based on the observed patterns, the next highest-value actions are:
\begin{enumerate}
\item native-speaker review of the highest edit-distance utterances rather than random examples alone;
\item speaker-, domain-, duration-, and scripted-status stratification;
\item explicit insertion, deletion, and substitution reporting;
\item a lexical normalization policy that distinguishes acceptable spelling variants from actual recognition errors;
\item a manually verified code-switch and named-entity subset;
\item endpointing and short-utterance challenge evaluations;
\item confidence calibration and selective prediction for cases in which the system should abstain or request repetition.
\end{enumerate}

\section{Prototype Streaming Deployment and Artifact Promotion}
\label{sec:deployment}

\subsection{True-Streaming Backend}

The deployed service restores the selected \texttt{.nemo} archive once per warm container, applies the \texttt{sw-KE} prompt, configures \texttt{[56,13]}, and exposes a stateful WebSocket stream.
Audio frames arriving from a browser are converted to the expected mono sample representation and appended to the active session.
The server retains model caches between messages and emits partial hypotheses without retranscribing the entire recording.

The isolated Kikuyu and Dholuo true-streaming services expose stateful
WebSocket interfaces for inference.

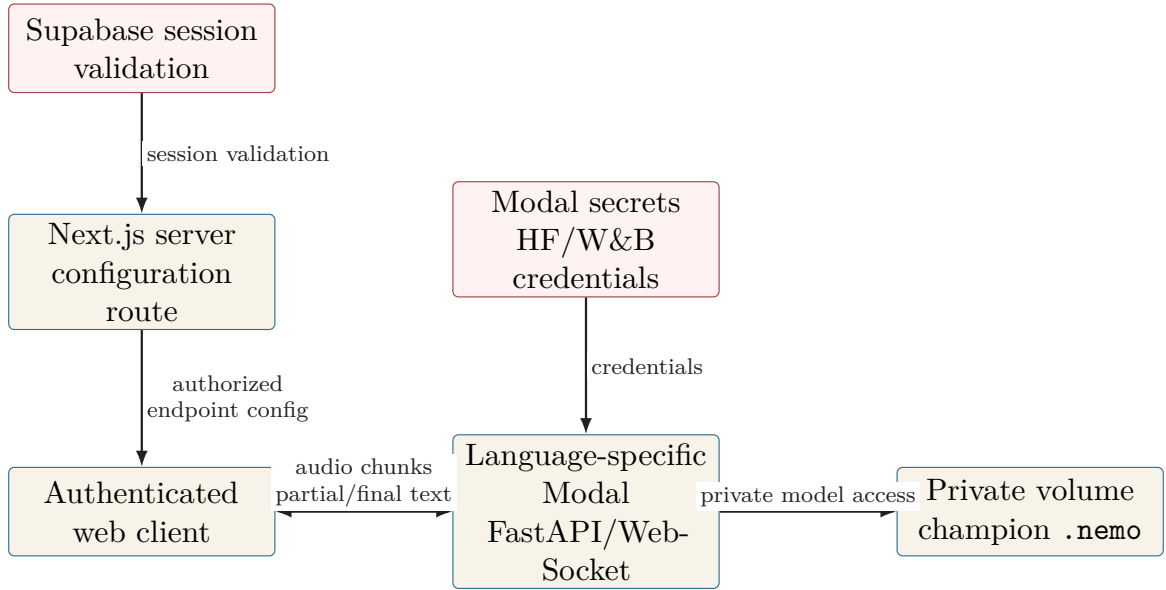
\begin{figure}[H]
\centering
\resizebox{0.96\textwidth}{!}{%
\begin{tikzpicture}[
  box/.style={
    draw=CeloBlue,
    rounded corners=2pt,
    align=center,
    minimum height=11mm,
    text width=31mm,
    inner sep=3pt,
    fill=CeloLight
  },
  secret/.style={
    draw=CeloRed,
    rounded corners=2pt,
    align=center,
    minimum height=11mm,
    text width=31mm,
    inner sep=3pt,
    fill=red!5
  },
  arrow/.style={-{Latex[length=2mm]},thick,CeloInk},
  stream/.style={{Latex[length=2mm]}-{Latex[length=2mm]},thick,CeloInk},
  flowlabel/.style={
    font=\scriptsize,
    align=center,
    fill=white,
    inner sep=1.5pt,
    text=CeloInk
  }
]

% Primary inference path
\node[box] (browser) {Authenticated\\web client};
\node[box,right=22mm of browser] (modal)
  {Language-specific\\Modal FastAPI/WebSocket};
\node[box,right=22mm of modal] (modelnode)
  {Private volume\\champion \texttt{.nemo}};

% Configuration and trust services
\node[box,above=17mm of browser] (route)
  {Next.js server\\configuration route};
\node[secret,above=15mm of route] (auth)
  {Supabase session\\validation};
\node[secret,above=17mm of modal] (secrets)
  {Modal secrets\\HF/W\&B credentials};

% Control and authorization flow
\draw[arrow] (auth) --
  node[flowlabel,right]{session validation}
  (route);

\draw[arrow] (route) --
  node[flowlabel,right]{authorized\\endpoint config}
  (browser);

% Live streaming flow
\draw[stream] (browser) --
  node[flowlabel,above]{audio chunks\\partial/final text}
  (modal);

% Server-side protected resources
\draw[arrow] (secrets) --
  node[flowlabel,right]{credentials}
  (modal);

\draw[arrow] (modal) --
  node[flowlabel,above]{private model access}
  (modelnode);

\end{tikzpicture}%
}

\caption{Prototype deployment and trust boundary. After server-side session verification, the configuration route returns the selected backend endpoint. The browser then opens a direct WebSocket connection to the language-specific service. The evaluated prototype did not implement a user-bound authorization token at the streaming-service ingress.}
\label{fig:deployment}
\end{figure}

\subsection{Language Isolation}

Kikuyu and Dholuo are deployed as separate Modal applications, profiles, volumes, model paths, and frontend configuration routes.
This isolation reduces blast radius: replacing the Dholuo champion cannot silently alter the Kikuyu service.
The user interface presents language selection; after server-side session verification, the configuration route returns the corresponding backend URL and the browser connects to that language service.
This application gate protects configuration discovery, but the backend streaming endpoint does not yet require a short-lived user-bound stream token.
Kalenjin remains a placeholder until it meets the project's release threshold.

\subsection{Authentication and API Protection}

The C-elo AI page uses Supabase Google OAuth.
Unauthenticated navigation is redirected to a sign-in route, and server-side API routes verify the Supabase session before returning model-service configuration.
An unauthenticated request to the configuration API returns HTTP 401.
This prevents the page itself from being the only access control.

Authentication does not make a browser-visible WebSocket URL secret.
Any URL used directly by browser JavaScript can be observed in developer tools.
Abuse resistance therefore requires authorization at the service boundary, short-lived signed access where appropriate, per-account quotas, concurrency limits, origin checks as defense in depth, request-size and duration limits, and server-side usage accounting.
These controls were not all part of the evaluated prototype: the implemented boundary validates Supabase sessions on protected application and configuration routes, while direct stream-token authorization, metering, and account-level quotas remain production-hardening work.

\subsection{Checkpoint Confidentiality}

The inference service mounts a private persistent volume and restores the model server-side.
Neither the \texttt{.nemo} file nor its filesystem path is returned to clients.
Private Hugging Face mirrors require an account token with repository permission.
The public application holds no Hugging Face write token, Modal credential, W\&B token, Supabase secret key, or model binary.

The Supabase publishable client key is intentionally browser-visible and is constrained by authentication and row-level security; privileged service-role or secret keys remain server-side.
Likewise, model-service credentials are stored as platform secrets, not \texttt{NEXT\_PUBLIC\_*} variables.

\subsection{Artifact Promotion}

The project defines the following target model-promotion workflow for new releases:
\begin{enumerate}
\item select the lowest-validation-WER checkpoint;
\item run the complete true-streaming internal evaluation;
\item compare against the current champion under the same manifest and normalization;
\item inspect qualitative and stratified regressions;
\item export a self-contained \texttt{.nemo} archive;
\item calculate checksums and create a machine-readable artifact manifest;
\item copy the champion into an immutable-named folder on the training volume;
\item mirror it to a private Hugging Face repository;
\item deploy to an isolated language backend;
\item perform application-gated live smoke tests before switching the frontend configuration.
\end{enumerate}

Kikuyu v13 followed the validation-selected export path.
Dholuo v7 is a historical exception: its final run export was evaluated and deployed, while the slightly better validation checkpoint was preserved separately.
Accordingly, the reported Dholuo test metrics characterize the final v7 \texttt{.nemo} export rather than a restored best-validation checkpoint.

No training run updates the live model automatically and Kalenjin has not been promoted to a live release yet.
The sequence above is therefore both a record of the Kikuyu and Dholuo promotion path and the standard for subsequent promotions.

\section{Reproducibility and Artifact Provenance}
\label{sec:reproducibility}

\subsection{What Is Preserved}

For each promoted model, the project internally records the model archive,
source run and timestamp, selected validation metric, internal evaluation
results, corpus-generation identifier, prompt, attention context, decoding
mode, software entrypoint, and model checksum. These records are distributed
across controlled Modal storage, private Hugging Face mirrors, experiment
blueprints, and evaluation logs; they are not publicly available as a
single complete artifact bundle.

The \texttt{.nemo} file is sufficient to restore the trained model under a
compatible NeMo environment because it contains the learned weights, model
configuration, and tokenizer assets. Additional manifests, resolved
configurations, evaluation reports, and checksums support provenance
auditing and evaluation reproduction. Trainer checkpoints provide the
additional state needed when continuation or training resumption is
required. External reproduction remains subject to the access restrictions
described in the Data and Artifact Availability statement.

\subsection{Reproducibility Levels}

This project distinguishes four levels:
\begin{description}[leftmargin=3.2cm,style=nextline]
\item[Artifact reproducibility] The exact exported model can be restored and its checksum verified.
\item[Evaluation reproducibility] The model, manifest, normalization, prompt, context, and decoder can be rerun to reproduce metrics within infrastructure tolerance.
\item[Training reproducibility] The data generation, initialization, hyperparameters, software environment, and run command are available to repeat fine-tuning, although stochastic kernels may prevent bitwise equality.
\item[Scientific reproducibility] An independent group can evaluate the method on accessible data and verify the conclusions without relying on private internal artifacts alone.
\end{description}

The present work is strongest at the first two levels inside the controlled project environment.
Full scientific reproducibility is constrained by dataset access terms, private champion repositories, platform-specific infrastructure, and the need to publish a frozen external evaluation protocol.

\subsection{Experiment Ledger}

The principal continuation-run names encode the language, stage, hardware, parent checkpoint, corpus size or generation, learning rate, step budget, and validation regime.
This verbose naming convention is intentional.
It makes accidental initialization from the wrong model or evaluation against the wrong corpus easier to detect.
The appendices provide the principal lineage and command templates; archived timestamps, resolved configurations, paths, blueprints, and artifact manifests supply details that are not present in every run name.

\section{Responsible Development}
\label{sec:responsible}

\subsection{Community and Language Expertise}

Orthographic and annotation decisions cannot be delegated entirely to Unicode rules or an English-language researcher.
Native-speaker review was used for Kikuyu character mappings and Kalenjin marker interpretation.
For Kalenjin, the speaker confirmed the linguistic content of the reviewed examples while identifying the exact forms \texttt{[cs]}, \texttt{(cs)}, \texttt{[pause]}, and \texttt{(pause)} as annotation wrappers rather than spoken target words.

This process should grow from ad hoc consultation into documented community governance.
Future releases should identify who defines acceptable orthography, how disagreements are adjudicated, how contributors are credited and compensated, and how users can report harmful or systematically incorrect outputs.

\subsection{Privacy and Consent}

Speech can reveal identity, location, health, and social relationships.
The use of African Next Voices data must remain consistent with the dataset license, consent model, and access conditions \cite{wanzare2026afrivoices,anv_kikuyu,anv_dholuo,anv_kalenjin}.
The browser and server were designed not to intentionally persist raw microphone audio beyond transient processing, and the interface tells users that audio is processed for transcription.
This study did not conduct an end-to-end audit of retention by every infrastructure component. A production release should therefore configure server-side logging to avoid retaining raw audio or sensitive transcripts unless an explicit, consented diagnostic program is active.

\subsection{Unequal Error Costs}

A single WER does not capture social harm.
Errors in casual dictation differ from errors in medical, legal, financial, or government-service contexts.
Named entities, numbers, medication terms, and negation can be high-consequence even when aggregate WER is unchanged.
The current models are research and testing systems, not certified decision-making tools.
High-stakes deployment requires domain-specific testing, calibrated uncertainty, human verification, and a clear refusal or escalation path.

\subsection{Language Variety and Representation}

Labels such as ``Kikuyu,'' ``Dholuo,'' and ``Kalenjin'' can hide dialect, region, age, gender, and code-switching variation.
The speaker and recorder counts reported during curation are useful but do not prove balanced representation.
Future dataset cards and model cards should provide subgroup coverage where consent and privacy permit, accompanied by stratified performance and known limitations.

\section{Limitations}
\label{sec:limitations}

The study has several important limitations.
\begin{enumerate}
\item \textbf{No paired external benchmark.} Internal results are not evaluated on the exact PazaBench or another common public test set, so state-of-the-art claims are not supported.
\item \textbf{Repeated held-out consultation.} Successive champion decisions used the internal test sets; they are not pristine one-shot evaluations.
\item \textbf{Kalenjin mixed-source selection.} The historical v1-v run ranked checkpoints with a validation manifest containing test-origin rows. Its validation and clean-v3 test scores are adaptive diagnostics, not independent generalization estimates.
\item \textbf{Different corpus generations.} Kalenjin clean-v1, clean-v2, and clean-v3 alter references and row membership, so not every apparent gain is attributable solely to learned weights.
\item \textbf{Incomplete Kalenjin metrics.} The v1-v evaluator archived WER but not CER, limiting error-type comparison with prior stages.
\item \textbf{No controlled seed study.} Runs are expensive single trajectories; variance across seeds is unknown.
\item \textbf{No tokenizer ablation.} The inherited 13,087-piece tokenizer is retained, and its effect on target-language morphology is not isolated.
\item \textbf{No optimizer-continuous comparison.} Staged \texttt{.nemo} continuation is not compared with a matched uninterrupted trainer-checkpoint resume.
\item \textbf{Limited latency characterization.} Evaluation runtime and one aggregate batched real-time factor are reported, but interactive production latency, concurrent-stream throughput, and cold starts require a dedicated benchmark.
\item \textbf{Greedy decoding only.} The report does not compare beam search, shallow fusion, rescoring, or target-language language models.
\item \textbf{No formal fairness audit.} Recorder and domain summaries exist, but comprehensive demographic subgroup analysis is not yet available.
\item \textbf{Private artifacts and access controls.} Some data and model artifacts are intentionally private, which limits immediate independent reproduction.
\end{enumerate}

These limitations do not invalidate the observed internal improvements.
They define the scope of the evidence and the experiments needed before stronger claims are made.

\section{Discussion}
\label{sec:discussion}

\subsection{What Drove Improvement}

The project does not support a single-cause narrative.
Three factors interact:
\begin{enumerate}
\item a strong multilingual streaming initialization with an architecture already suited to incremental decoding;
\item a Kenyan Swahili-adapted initialization that plausibly reduced adaptation distance in the project setting, although its advantage over the raw NVIDIA base was not isolated;
\item repeated data and evaluation audits that remove supervision artifacts and ensure the tested path matches deployment.
\end{enumerate}

The learning curves show that continued full-parameter adaptation remains effective at a small learning rate.
The corpus revisions show that a better metric can also result from a more valid reference.
Scientific reporting must keep these distinct: parameter improvement, label correction, and evaluation-protocol change are different interventions.

\subsection{Why a Streaming Base Matters}

An offline encoder-decoder can be adapted for chunked use, but the resulting engineering does not automatically preserve the behavior measured offline.
Starting with a cache-aware streaming model allows training, trainer
validation, true-streaming evaluation, and deployment to share the same
model family, configured attention context, and decoder family while using
different execution paths.
This reduces the gap between a research checkpoint and an interactive product.

Whisper remains an important general-purpose baseline \cite{radford2023whisper}, but its standard sequence-to-sequence transcription interface and long-window decoding assumptions differ from the stateful RNN-T path used here.
The relevant comparison is not that one family is universally superior; it is whether the model's architectural and deployment assumptions match the required interaction.

\subsection{Data Quality Versus Data Quantity}

The Kalenjin clean-v3 corpus is smaller than clean-v1, yet it provides a cleaner experimental target.
Dropping rows is not automatically beneficial: excessive filtering can erase dialects, rare terms, code-switching, and hard acoustic conditions.
The key is whether a row is hard but valid or invalid supervision.
Hard valid examples should be preserved and weighted appropriately; invalid examples should be corrected or excluded.

This distinction motivates future confidence-based curation with native-speaker adjudication rather than increasingly aggressive regular-expression filters.

\subsection{Implications for Other Low-Resource Languages}

The transferable lesson is a workflow rather than a single checkpoint.
Begin from a deployment-compatible multilingual model; choose a related bridge only when its acoustic and linguistic rationale is explicit; version every corpus transformation; validate audio and text jointly; use full validation for checkpoint selection; evaluate through the intended streaming path; and preserve enough provenance to reverse decisions.

This approach is applicable beyond the three languages studied, but the exact character maps, marker policies, and acceptable variants must be determined language by language.

\section{Future Work}
\label{sec:future}

The immediate research agenda is:
\begin{enumerate}
\item freeze a new speaker-disjoint one-shot benchmark for Kikuyu, Dholuo, and Kalenjin;
\item reproduce at least one public benchmark exactly, including normalization and decoding rules;
\item add standard CER and no-space CER to every Kalenjin evaluator output;
\item conduct native-speaker adjudication of the most frequent Kalenjin boundary and lexical errors;
\item compare inherited and target-aware tokenizers without changing the acoustic initialization;
\item evaluate beam search, external language-model fusion, and contextual biasing for names and service terms;
\item study short-utterance endpointing with manually verified boundaries;
\item report multiple random seeds or bootstrap confidence intervals;
\item benchmark cold start, real-time factor, first-token delay, finalization delay, and concurrent sessions;
\item quantify performance by recorder, scripted status, domain, duration, gender, age, and dialect where ethically and statistically valid;
\item build model cards and dataset cards with community review, intended-use boundaries, and incident-reporting channels;
\item evaluate whether parameter-efficient adaptation can approach full fine-tuning while reducing per-language storage and compute.
\end{enumerate}

\section{Conclusion}
\label{sec:conclusion}

This work demonstrates an end-to-end route from a multilingual streaming ASR foundation model to functioning Kenyan-language speech systems.
The contribution is not only a pair of promoted champion model artifacts.
It is the combination of data forensics, bridge initialization, controlled full-parameter adaptation, true-stream evaluation, explicit checkpoint promotion, isolated deployment, and preservation.

Under the internal adaptively consulted true-streaming protocols, the validation-selected Kikuyu model reaches 42.9677\% WER and 7.7947\% no-space CER, while the evaluated Dholuo v7 final export reaches 33.9791\% WER, 9.5855\% standard CER, and 8.1340\% no-space CER under its frozen historical \texttt{cs} label policy.
The Kalenjin v1-v checkpoint produced 68.74\% WER on the filtered clean-v3 adaptive diagnostic, but mixed-source checkpoint selection and repeated consultation preclude treating that value as an independent generalization estimate; Kalenjin remains research in progress.
These numbers show substantial learning within the project; they are not presented as externally verified state of the art.

The broader finding is that low-resource ASR is as much a problem of valid supervision, evaluation discipline, and deployment fidelity as it is a problem of model scale.
When those pieces are treated as one auditable, versioned system, PyTorch and NeMo make it possible to move from research experiments to interactive speech technology without abandoning scientific caution.

\appendix

\section{Corpus Construction and Normalization Specification}
\label{app:data-specification}

This appendix records the operational definition of the corpora used in the experiments.
It is intentionally more detailed than a conventional conference methods section because corpus construction was one of the principal research variables.
The source datasets contain valid difficult speech, transcription conventions, unavailable audio objects, normalization artifacts, and split metadata that cannot safely be treated as interchangeable.

\subsection{Manifest Contract}

Every accepted example is represented as one JSON object per line with at least the following fields:

\begin{lstlisting}[language=json,caption={Canonical NeMo manifest record used by the project.}]
{
  "audio_filepath": "/data/.../audio/train/00001234.wav",
  "duration": 7.42,
  "text": "normalized target language transcription"
}
\end{lstlisting}

Additional provenance fields are retained when available, including source split, source row identifier, recording type, domain, recorder identifier, original filename, and pre-normalization text.
Training consumes the three canonical fields, while the additional fields support auditing and stratified analysis.

An accepted row must satisfy all of the following invariants:
\begin{enumerate}
  \item the referenced audio object exists and decodes successfully;
  \item duration is finite, positive, and no greater than the maximum configured for the corresponding language and corpus generation;
  \item normalized text is non-empty and contains no Unicode replacement character;
  \item the text conforms to the language-specific normalization policy;
  \item the row belongs to exactly one model-selection role: training, validation, or held-out test;
  \item the chosen split key does not overlap another role after canonicalization;
  \item the manifest is valid JSONL and can be scanned from beginning to end without a parse error.
\end{enumerate}

\subsection{Common Audio Pipeline}

The common audio path is summarized in Listing~\ref{alg:common-preparation}.
The implementation uses structured dataset records and audio decoders rather than inferring metadata from filenames alone.

\begin{lstlisting}[caption={Language-independent corpus preparation pseudocode.},label={alg:common-preparation}]
max_duration = duration_limit_seconds[language]

for source_row in source_split:
    raw_text = read_transcription(source_row)
    if raw_text is missing or blank:
        reject("empty_transcription")

    normalized = normalize_unicode_and_text(raw_text, language)
    if normalized is blank:
        reject("empty_after_normalization")
    if contains_unicode_replacement_character(normalized):
        reject("replacement_character")

    audio = resolve_audio_object(source_row)
    try:
        waveform, sample_rate = decode(audio)
    except DecodeError:
        reject("decode_error")

    duration = num_samples(waveform) / sample_rate
    if duration <= 0 or duration > max_duration:
        reject("invalid_or_too_long")

    waveform_16k = resample_mono(waveform, sample_rate, 16000)
    output_path = deterministic_output_path(source_row)
    write_audio(output_path, waveform_16k, 16000)
    emit_manifest_record(output_path, duration, normalized, provenance)

verify_jsonl()
verify_audio_references()
verify_split_disjointness()
write_summary_and_checksums()
\end{lstlisting}

Audio is materialized into a persistent Modal volume so that training runs do not repeatedly depend on remote Parquet delivery.
This design also isolates model training from transient network failures encountered while reading the source repository.
Remote reads that raised incomplete-transfer errors were retried; rows that still lacked a decodable object were excluded and counted rather than silently represented by an empty waveform.

\subsection{Common Text Operations}

The normalization function applies a deterministic ordered sequence:
\begin{enumerate}
  \item normalize Unicode representation to a stable form;
  \item standardize apostrophe and quotation variants where their linguistic role is known;
  \item lowercase according to the project transcription convention;
  \item normalize whitespace;
  \item remove punctuation that is outside the target transcription inventory;
  \item apply a language-specific character map only when the mapping is unambiguous;
  \item remove or reject verified annotation markers according to the corpus version;
  \item reject the row if uncertainty cannot be resolved without inventing lexical content.
\end{enumerate}

The final operation is important.
Normalization is not a license to guess what a damaged symbol was intended to mean.
Rows with uncertain character recovery are rejected so that the model is not trained to imitate a fabricated target.

\subsection{Kikuyu Orthography Policy}

Kikuyu required explicit preservation of the orthographic vowels \textit{\~{\i}} and \textit{\~{u}} and their uppercase forms before lowercasing.
Source records also contained visually similar or decomposed forms introduced by keyboards, exports, and encoding conversions.
The pipeline canonicalized only verified equivalents, retained apostrophes when linguistically meaningful, and rejected rows containing an unresolved replacement character.

The quality-control question was not merely whether a string was valid Unicode.
It was whether two strings represented the same intended Kikuyu transcription.
This distinction prevented broad ASCII folding, which would have erased contrasts that matter to both lexical identity and evaluation.

\subsection{Dholuo Text Policy}

Dholuo uses a predominantly Latin orthography but includes
apostrophe-bearing forms and extensive code-switching in the source
material. The historical pipeline standardized quotes, spacing, casing,
and punctuation while retaining lexical apostrophes. It did not include a
rule for removing the literal token \texttt{cs}; validation logs confirm
that the token remained in references after normalization.

A fixed but imperfect label policy can support controlled comparisons
among checkpoints evaluated on the same split and with the same scoring
configuration when that policy is disclosed and applied consistently.
It does not support direct comparison with a corpus or benchmark using a
different label or normalization policy.

\subsection{Kalenjin Cleaning Generations}

Kalenjin was deliberately versioned through three cleaning generations instead of overwriting a single manifest.

\paragraph{clean-v1.}
The first complete corpus applied the common decoding, duration, Unicode, and text-normalization rules.
It exposed two additional problems during evaluation: non-lexical annotation markers embedded in otherwise valid text, and short or artifact-heavy labels that encouraged fluent over-generation.

\paragraph{clean-v2.}
A Kalenjin speaker reviewed examples and confirmed that the exact markers \texttt{[cs]}, \texttt{(cs)}, \texttt{[pause]}, and \texttt{(pause)} were annotations rather than words to be recognized.
clean-v2 removes those exact markers while preserving surrounding lexical material.
The policy does not blindly remove all bracketed content because a bracket could contain meaningful text in another corpus.

\paragraph{clean-v3.}
The third generation derives a stricter corpus from clean-v2.
It:
\begin{enumerate}
  \item retains the four confirmed-marker removals from clean-v2;
  \item drops an entire row if its normalized reference contains the literal annotation phrase \texttt{long pause};
  \item drops an entire row if it contains a digit or a mixed alphanumeric artifact such as \texttt{20082018} or \texttt{we5107};
  \item drops a row with fewer than three normalized tokens to reduce the most severe short-utterance over-generation cases.
\end{enumerate}

The digit rule rejects the row; it does not simply delete the offending token.
Deleting a token from an otherwise time-aligned transcription can make the remaining label falsely claim that audible content never occurred.
Whole-row rejection sacrifices a small amount of audio but avoids introducing a known source of audio--text misalignment into the accepted set.

\begin{lstlisting}[caption={Kalenjin clean-v3 derivation pseudocode.}]
for row in clean_v2_manifest:
    text = normalize_whitespace(row.text)
    text = remove_exact_markers(
        text, ["[cs]", "(cs)", "[pause]", "(pause)"])

    if contains_literal_phrase(text, "long pause"):
        reject_entire_row("long_pause_marker")
    if contains_any_digit(text):
        reject_entire_row("digit_or_alphanumeric_artifact")
    if token_count(text) < 3:
        reject_entire_row("too_short")

    emit(row.with_text(normalize_whitespace(text)))
\end{lstlisting}

\subsection{Why Valid Hard Examples Were Not Broadly Removed}

The cleaning rules are deliberately narrow.
Accents, dialectal pronunciation, code-switching, background noise, rare names, and long morphologically complex words can all increase error while remaining valid supervision.
Removing such examples simply because the current model finds them difficult would bias the corpus toward an artificially easy language variety.
The project therefore distinguishes:
\begin{itemize}
  \item \textbf{hard-valid}: difficult speech whose audio and reference plausibly agree;
  \item \textbf{invalid-supervision}: missing audio, corrupt text, verified non-speech annotations represented as words, unresolved encoding damage, or a reference whose alignment has been invalidated; and
  \item \textbf{scope-defining experimental exclusion}: potentially valid rows, including digit-bearing or fewer-than-three-token references, excluded specifically to define the clean-v3 experimental population.
\end{itemize}

Only invalid supervision is a routine data-quality rejection target.
Scope-defining exclusions apply only to the stated experimental generation and must be reported with every result computed on it.
Speaker adjudication is required when the category is uncertain.

\section{Detailed Corpus Ledgers}
\label{app:corpus-ledgers}

\subsection{Dholuo Full-Corpus Preparation}

The source repository exposed 117,564 training rows and 13,853 validation rows at preparation time.
The deterministic preparation summary is shown in Table~\ref{tab:dholuo-ledger}.

\begin{table}[H]
\centering
\caption{Dholuo preparation ledger. Hours are decoded accepted audio hours.}
\label{tab:dholuo-ledger}
\begin{tabular}{lrr}
\toprule
Item & Train & Validation \\
\midrule
Rows seen & 117,564 & 13,853 \\
Rows written & 92,287 & 11,152 \\
Hours retained & 206.8679 & 25.1899 \\
Replacement-character rejects & 1,913 & 221 \\
Decode-error rejects & 41 & 3 \\
Duration-over-limit rejects & 11,159 & 1,138 \\
Empty-transcription rejects & 12,164 & 1,339 \\
Scripted accepted rows & 85,394 & 10,271 \\
Unscripted accepted rows & 6,893 & 881 \\
Unique recorders & 661 & 87 \\
Rows whose text changed in normalization & 95,049 & 11,129 \\
Raw rows containing digits & 3,698 & 397 \\
Raw rows containing smart quotes & 4,944 & 548 \\
\bottomrule
\end{tabular}
\end{table}

The final Dholuo held-out evaluation manifest used after the split audit contains 5,480 rows.
Its role is separate from the full validation set used to monitor training.
For v7, the experiment ledger records a best validation WER of 0.33944 and a final logged validation WER of 0.33949.
The reported internal-evaluation WER and CER values apply to the final exported \texttt{.nemo} model; they should not be interpreted as measurements of the separately preserved best-validation trainer checkpoint.

The source domain distribution is not uniform.
Most accepted scripted rows have no explicit domain label, while the unscripted tail includes agriculture and food, everyday scenarios, financial transactions, named entities, role play, education and technology, healthcare, digital government services, stories, news, and customer care.
This imbalance motivates per-domain reporting in future work rather than interpreting a single aggregate WER as uniform language coverage.

\subsection{Kalenjin clean-v1 Ledger}

\begin{table}[H]
\centering
\caption{Kalenjin clean-v1 preparation ledger.}
\label{tab:kalenjin-v1-ledger}
\begin{tabular}{lrrr}
\toprule
Item & Train & Validation & Test \\
\midrule
Rows seen & 82,378 & 13,845 & 3,315 \\
Rows written & 68,012 & 12,084 & 2,557 \\
Hours retained & 158.9356 & 25.5004 & 5.6431 \\
Replacement-character rejects & 450 & 27 & 11 \\
Decode-error rejects & 392 & 4 & 2 \\
Duration-over-limit rejects & 12,621 & 1,572 & 629 \\
Empty-transcription rejects & 902 & 158 & 116 \\
Empty-after-normalization rejects & 1 & 0 & 0 \\
Scripted accepted rows & 58,927 & 11,130 & 2,389 \\
Unscripted accepted rows & 9,085 & 954 & 168 \\
Unique recorders & 361 & 46 & 23 \\
\bottomrule
\end{tabular}
\end{table}

The large training rejection count is primarily explained by the fixed 39.99-second cap, not by wholesale linguistic exclusion.
The accepted training set retains both scripted and unscripted material and spans everyday scenarios, news and media, healthcare, agriculture and food, education and technology, finance, customer care, role play, extempore stories, named entities, and digital-government speech.

\subsection{Kalenjin clean-v3 Derivation Ledger}

\begin{table}[H]
\centering
\caption{Kalenjin clean-v3 derivation from clean-v2.}
\label{tab:kalenjin-v3-ledger}
\begin{tabular}{lrrrrrrr}
\toprule
Split & Read & Kept & Dropped & Hours & Digit & Short & Long pause \\
\midrule
Train & 68,012 & 64,007 & 4,005 & 151.28 & 2,539 & 1,463 & 3 \\
Validation & 12,084 & 11,364 & 720 & 24.25 & 358 & 357 & 5 \\
Test & 2,557 & 2,411 & 146 & 5.33 & 99 & 46 & 1 \\
\bottomrule
\end{tabular}
\end{table}

The model-selection validation file is a speaker/source-audited subset of the derived validation material:
\begin{itemize}
  \item training: 64,007 rows, 151.28 hours;
  \item \texttt{validation\_dev.json}: 8,953 rows, 18.92 hours;
  \item held-out \texttt{test.json}: 2,411 rows, 5.33 hours.
\end{itemize}

The raw derived \texttt{validation.json} still includes rows whose source role is \texttt{dev\_test}; therefore it must not be used directly for model selection.
The verification procedure reported zero JSON errors, missing audio objects, empty labels, replacement characters, train--validation overlap, train--test overlap, and validation--test overlap for the three role-specific manifests above.

\subsection{Kikuyu Curated Corpus Summary}

The final Kikuyu continuation series uses the curated approximately 182-hour training corpus established in the project ledger.
The curation emphasizes canonical Kikuyu orthography and removal of unresolved encoding corruption.
The final continuation series selected checkpoints on the filtered-v4
\texttt{validation\_full.json}, containing 13,667 utterances
(19.64 hours).
Because multiple historical source and curation generations preceded the final 182-hour corpus, this paper reports the frozen corpus statistics, manifest roles, and evaluation protocol used by the final model rather than reconstructing counts from experiment names. Raw ANV audio, speaker metadata, and other source-controlled materials are not redistributed; access remains subject to the source-provider terms described in the Data and Artifact Availability statement.

\section{Model Configuration Ledger}
\label{app:model-ledger}

\subsection{Restored Foundation Configuration}

The experiments restore the complete NeMo archive and then override only the intended data, optimization, validation, checkpointing, prompt, and streaming-decoding settings.
The restored model contains the following principal components:
\begin{itemize}
  \item 24 cache-aware FastConformer encoder layers;
  \item encoder hidden dimension 1024;
  \item a 128-dimensional one-hot language identifier that is broadcast across encoder time and fused with the acoustic representation;
  \item a 13,087-entry SentencePiece vocabulary inherited from the multilingual model;
  \item an RNN-T prediction network and joint network;
  \item prompt-conditioned language control;
  \item approximately 637 million trainable parameters in the restored project configuration;
  \item cache-aware streaming attention with the selected context \texttt{[56,13]}.
\end{itemize}

The first context number describes the retained left-context capacity in the NeMo attention configuration and the second controls right look-ahead.
The notation is a model configuration, not a number of seconds.
The effective temporal behavior depends on the feature hop, subsampling, convolutional context, cache policy, and chunking logic.

\subsection{Recorded Fine-Tuning Configuration}

The core continuation configuration and the explicitly recorded
stage-specific differences are summarized in
Table~\ref{tab:shared-hparams}. The principal selected branches generally
used full-parameter AdamW continuation with a learning rate of
\(5\times10^{-7}\), duration-based training and validation batches, and
validation every 1,000 optimizer steps. Warm-up and checkpoint-retention
settings were stage-specific rather than universal; consequently, this
appendix does not infer one shared value for every experiment.

The fixed small learning rate is a conservative continuation choice.
It avoids a large optimizer shock when moving from one language-adapted checkpoint to the next, but it is not claimed to be globally optimal.
The fixed learning rate was associated with stable incremental improvements across the observed continuation lineages, but these single-lineage experiments do not establish that it is optimal or that the same behavior will generalize to other datasets or initializations.
Future work should compare decayed schedules and multiple seeds under a fixed compute budget.

\subsection{Training Objective}

For acoustic sequence $\mathbf{x}$ and target token sequence $\mathbf{y}$, RNN-T marginalizes all monotonic alignments $\pi$ that collapse to $\mathbf{y}$:
\begin{equation}
  \mathcal{L}_{\mathrm{RNNT}}(\mathbf{x},\mathbf{y},\mathbf{q}_{\ell})
  = -\log \sum_{\pi \in \mathcal{B}^{-1}(\mathbf{y})}
  P(\pi \mid \mathbf{x},\mathbf{q}_{\ell}).
\end{equation}

For selected language identifier $\ell$, the model constructs a one-hot vector
\begin{equation}
  \mathbf{q}_{\ell}=\operatorname{onehot}(\ell)\in\mathbb{R}^{128}.
\end{equation}
The vector is broadcast across encoder time, concatenated with the acoustic representation, and passed through the prompt-fusion projection:
\begin{equation}
  \widetilde{\mathbf{h}}^{\mathrm{enc}}_t
  = f_{\mathrm{prompt}}\!\left(
      [\mathbf{h}^{\mathrm{enc}}_t;\mathbf{q}_{\ell}]
    \right).
\end{equation}
The prediction network maps non-blank label history to $\mathbf{h}^{\mathrm{pred}}_u$, and the joint network produces logits
\begin{equation}
  \mathbf{z}_{t,u}
  = f_{\mathrm{joint}}(\widetilde{\mathbf{h}}^{\mathrm{enc}}_t,
                       \mathbf{h}^{\mathrm{pred}}_u).
\end{equation}
FastEmit regularization encourages earlier non-blank emissions without changing the reference sequence.

\section{Complete Experiment Registry}
\label{app:experiment-registry}

This appendix separates the chronological research path from the compact result tables in the main text.
Run names are preserved because they encode source checkpoint, corpus size, learning rate, step budget, and validation mode.

\subsection{Kikuyu Continuation Series}

\small
\begin{longtable}{L{1.0cm}L{3.2cm}C{1.2cm}C{1.2cm}L{6.6cm}}
\caption{Kikuyu experiment registry. WER and NS-CER are percentages on the corresponding adaptively consulted internal evaluation set.}\label{tab:kikuyu-registry}\\
\toprule
Stage & Initialization / intervention & WER & NS-CER & Interpretation \\
\midrule
\endfirsthead
\toprule
Stage & Initialization / intervention & WER & NS-CER & Interpretation \\
\midrule
\endhead
v4 & Early language continuation & 53.3505 & 10.7256 & Established that the Swahili bridge could transfer to Kikuyu. \\
v6 & Continued low-rate full tuning & 51.2816 & 10.1190 & Reduced both boundary and character errors. \\
v7-i & Same model family and curated corpus & 50.8278 & 9.9395 & Incremental improvement; no architecture change required. \\
v8-ii & Further continuation & 49.3441 & 9.5613 & Passed below 50\% WER under the internal protocol. \\
v9 & Further continuation & 48.5926 & 9.3104 & Character quality continued to improve. \\
v10 & Curated approximately 182-hour path & 46.8393 & 8.7815 & Data curation and continuation jointly improved the system. \\
v11 & Same curated path & 45.6926 & 8.4489 & Stable continued learning. \\
v12 & 50,000-step continuation from v11 & 44.6985 & 8.1824 & Became the live champion before v13. \\
v13 & 70,000-step continuation from v12 & 42.9677 & 7.7947 & Current preserved Kikuyu champion; best validation WER 0.41163. \\
\bottomrule
\end{longtable}
\normalsize

The v13 final validation point was 0.41244 while an earlier retained checkpoint reached 0.41163.
This is why the project promotes a validation-selected checkpoint rather than assuming the final optimizer state is best.
In the Kikuyu v4 filtered experiment, the selected individual checkpoint obtained 53.3505\% WER and 10.7256\% NS-CER, whereas averaging the top three v4 checkpoints obtained 53.3887\% WER and 10.7683\% NS-CER under the same filtered full-test protocol.
The averaged candidate therefore did not improve the selected individual checkpoint; the available evidence does not isolate the cause.
For v13, the project exported the validation-selected best individual checkpoint, denoted \emph{best1}; this paper does not claim that a multi-checkpoint v13 average was evaluated.

\subsection{Dholuo Full-Data Series}

\small
\begin{longtable}{L{0.9cm}L{6.0cm}C{2.0cm}L{4.5cm}}
\caption{Dholuo experiment registry.}\label{tab:dholuo-registry}\\
\toprule
Stage & Training intervention & Best validation WER & Interpretation \\
\midrule
\endfirsthead
\toprule
Stage & Training intervention & Best validation WER & Interpretation \\
\midrule
\endhead
v3-i & 20,000 steps from the best earlier bridge-adapted Dholuo checkpoint on the 206.87-hour full training manifest & 0.49586 & First complete full-data continuation; confirmed the corpus and H200 path. \\
v4-i & 70,000 steps from v3-i, same corpus and $5\times10^{-7}$ rate & 0.38395 & Large absolute reduction with a smoothly descending validation curve. \\
v5-i & 40,000 steps from v4-i & 0.36139 & Continued improvement with unchanged optimization. \\
v6-i & 20,000 steps from v5-i & 0.35174 & Held-out true-stream WER 35.3502\%, NS-CER 8.3031\%. \\
v7-i & 40,000 steps from v6-i & 0.33944 & Current champion: internal-evaluation WER 33.9791\%, standard CER 9.5855\%, NS-CER 8.1340\%. \\
\bottomrule
\end{longtable}
\normalsize

The v7 final logged validation point was 0.33949 and the best retained value was 0.33944.
The held-out true-stream evaluation processed 5,480 rows in 222.45 seconds in the recorded run.
The source checkpoint and all retained evidence were copied into a separate champion directory and mirrored to a private Hugging Face repository with checksums and a model card.

\subsection{Kalenjin Research Series}

\begin{longtable}{L{1.15cm}L{1.35cm}r r r r L{4.15cm}}
\caption{Kalenjin experiment registry. Dashes indicate metrics not emitted by that evaluator. The v1-v validation and test values are adaptive diagnostics because its selection manifest contained test-origin rows.}\label{tab:kalenjin-registry}\\
\toprule
Stage & Labels & Val. WER & Test WER & CER & NS-CER & Interpretation \\
\midrule
\endfirsthead
\toprule
Stage & Labels & Val. WER & Test WER & CER & NS-CER & Interpretation \\
\midrule
\endhead
v1-i & clean-v1 & 0.83147 & 85.78 & -- & -- & First complete 20,000-step baseline from the Swahili bridge. \\
v1-ii & clean-v1 & 0.75607 & 78.0242 & 24.4985 & 22.8378 & 40,000-step continuation; strong relative improvement. \\
v1-iii & clean-v1 & 0.70793 & 73.2007 & 21.9547 & 20.6846 & 70,000-step continuation; exposed persistent marker and boundary errors. \\
v1-iii re-eval. & clean-v2 & 0.70793 & 73.0477 & 21.6998 & 20.4843 & Same weights, corrected labels; isolates a small label-policy effect. \\
v1-iv & clean-v2 & 0.68255 & 70.7831 & 20.4366 & 19.4145 & 70,000-step continuation on confirmed marker cleanup. \\
v1-v & clean-v3 & 0.66772 & 68.74 & -- & -- & 70,000-step continuation on clean-v3; mixed-source selection and adaptive diagnostic test. \\
\bottomrule
\end{longtable}

The reported diagnostic trend decreases across stages, but corpus revisions and the v1-v selection defect prevent attributing every delta solely to model learning; the series does not yet justify production status.
The archived v1-iv evaluator report records conventional CER 20.4366\% and NS-CER 19.4145\% on 2,557 clean-v2 test examples.
The clean-v3 v1-v adaptive diagnostic contains 2,411 examples (5.33 hours) and completed its streaming pass in 92.16 seconds.
The corresponding evaluator log did not emit CER or NS-CER, so those cells are not inferred from WER.

\subsection{Interpreting Continuation Stages}

Two consecutive 20,000-step jobs are generally not equivalent to one 40,000-step job.
A restart can reconstruct the optimizer, scheduler, random sampler state, gradient-scaler state, and data position differently.
It also changes checkpoint-selection opportunities and may change the source archive itself if the best intermediate model rather than the exact final state is used.
The project therefore treats every continuation as a distinct experiment with explicit initialization, not as an arithmetic addition of steps.

\section{Checkpoint Selection, Promotion, and Preservation}
\label{app:checkpoint-promotion}

\subsection{Selection Rule}

Checkpoint-retention depth was configured per run rather than fixed globally.
The selected late-stage settings are summarized in
Table~\ref{tab:shared-hparams}; earlier exploratory runs sometimes retained
fewer candidates.
The canonical release-selection workflow is:
\begin{enumerate}
  \item confirm training reached the requested optimizer-step budget;
  \item enumerate retained checkpoints and their validation metrics;
  \item select the lowest valid full-validation WER, not the newest file;
  \item restore the selected checkpoint into the exact model class;
  \item export a standalone \texttt{.nemo} archive;
  \item evaluate that archive on the held-out split through true streaming;
  \item compare it against the incumbent champion using the same label version, context, decoder, and normalization;
  \item promote only if the complete evidence package supports the change.
\end{enumerate}

Kikuyu v13 followed this canonical workflow.
For Dholuo v7, the final run \texttt{.nemo} export was evaluated and promoted; the best-validation trainer checkpoint was copied into the preservation package as separate evidence.

\subsection{Why a \texttt{.nemo} Archive Is the Deployment Unit}

A NeMo archive packages model configuration, learned weights, tokenizer artifacts, and other restoration assets required by the model class.
For the live service, the promoted \texttt{.nemo} file is the principal model artifact.
Training \texttt{.ckpt} files are larger because they can also contain optimizer and trainer state needed for resumption.
Evaluation reports, manifests, checksums, and source checkpoints remain valuable for provenance but are not all required merely to load the deployed recognizer.

\subsection{Target Champion Evidence Contract}

For future promotions, the project defines a target immutable
champion-evidence directory containing:
\begin{itemize}
  \item the canonical \texttt{model/<champion-name>.nemo};
  \item \texttt{champion\_manifest.json} or \texttt{artifact\_manifest.json};
  \item SHA-256 checksums for every preserved artifact;
  \item the full internal true-streaming evaluation report and evaluator configuration;
  \item source run identifier, source checkpoint path, W\&B run URL, and timestamp;
  \item dataset and label-version identifiers;
  \item prompt, context, decoder, and normalization settings;
  \item a human-readable model card with intended use and limitations.
\end{itemize}

\subsection{Integrity Verification}

For the preserved champion transfers used in the final deployments, the project computed SHA-256 digests, and the cross-profile download helpers rejected a \texttt{.nemo} file whose local digest did not match the expected champion digest.
File size is a useful first check but is not sufficient: two files can have the same size and different content.
The remote repository remains private, and the authentication token is supplied as a server-side secret rather than embedded in source code or a browser bundle.

\section{Reproducibility Recipes}
\label{app:reproducibility-recipes}

This appendix records the operational form of the principal jobs rather than only describing them in prose.
The commands are intentionally explicit about the profile, data-generation identifier, source archive, prompt, and streaming context.
They should be read as templates: run identifiers and timestamped checkpoint directories are immutable experiment inputs and must not be silently replaced by ``latest.''

\subsection{Environment and Secret Boundary}

The local machine contains the Modal client and the repository source, but it does not download the multi-gigabyte checkpoints for routine training.
The Modal application mounts the project source into a container and mounts a persistent volume at \texttt{/data}.
Authentication tokens for Hugging Face and W\&B are injected from Modal secrets.
At minimum, a reproducible environment records:

\begin{itemize}
  \item source revision or immutable source snapshot;
  \item Modal workspace/profile and volume name;
  \item container image digest or complete image build declaration;
  \item CUDA, PyTorch, NeMo, Python, and tokenizer versions;
  \item GPU type and count;
  \item training and validation manifest hashes;
  \item initialization-archive hash;
  \item every command-line override;
  \item W\&B run identifier and output directory.
\end{itemize}

The \texttt{/data} prefix is a container mount point, not part of the Modal volume namespace.
For example, the path \texttt{/data/checkpoints/x} inside a function is listed with
\texttt{modal volume ls <volume> /checkpoints/x} from the local CLI.
Confusing these namespaces produces a misleading ``No such file or directory'' error even when the object is present.

\subsection{Recommended Preflight Invariants}

The historical launchers checked required paths during job startup, but they did not enforce every invariant below in one pre-allocation stage.
The operational failures in Appendix~\ref{app:failure-ledger} motivate the following hardened launcher contract for future runs:

\begin{enumerate}
  \item training and validation manifests exist and are non-empty;
  \item all required JSON fields are present and parsable;
  \item sampled audio paths exist beneath the mounted volume;
  \item sampled durations are finite, positive, and within policy;
  \item normalized text is non-empty and contains no forbidden markers;
  \item the source \texttt{.nemo} archive exists and can be restored;
  \item the requested scheduler floor does not exceed the peak learning rate;
  \item the W\&B project and run name identify a new, intentional experiment;
  \item no function call for the same run name is already active.
\end{enumerate}

A hardened preflight should not merely count lines. It should compute a compact manifest fingerprint comprising row count, duration sum, text-token count, unique audio-path count, and SHA-256 digest.
This catches accidental use of a previous label generation even when filenames such as \texttt{train.json} are unchanged.

\subsection{Detached Continuation Training}

The following command captures the v1-v Kalenjin continuation pattern used after the clean-v3 corpus was created.
The important lifecycle property is \texttt{modal run -{}-detach}: the remote input survives closure of the local entrypoint and temporary loss of the laptop connection.

In the commands below, the profile name \texttt{c-elo} is a local CLI alias for the private Modal
workspace used in these experiments; reproducing the commands requires an
authorized profile with access to equivalent volumes, secrets, and checkpoints.

\begin{lstlisting}[caption={Detached full-parameter continuation on an H200.},label={lst:kalenjin-train}]
MODAL_PROFILE=c-elo KALENJIN_INCLUDE_PREMIUM_GPUS=1 \
modal run --detach \
  training/nemotron-asr-kalenjin/modal_train.py \
  --action smoke_train_h200 \
  --run-name \
    kalenjin_v1_v_h200_from_v1iv_full159h_cleanv3_lr5e7_70k_cleanval \
  --manifest-run-name kalenjin_full_clean_v3 \
  --train-manifest-name train.json \
  --validation-manifest-name validation.json \
  --max-steps 70000 \
  --learning-rate 5e-7 \
  --scheduler-min-lr 5e-7 \
  --batch-duration 200 \
  --validation-batch-duration 120 \
  --val-check-interval 1000 \
  --warmup-steps 100 \
  --save-top-k 10 \
  --target-lang sw-KE \
  --init-model-path \
    "/data/checkpoints/kalenjin_v1_iv_h200_from_v1iii_full159h_cleanv2_lr5e7_70k_cleanval/kalenjin_v1_iv_h200_from_v1iii_full159h_cleanv2_lr5e7_70k_cleanval/2026-06-29_21-55-52/checkpoints/kalenjin_v1_iv_h200_from_v1iii_full159h_cleanv2_lr5e7_70k_cleanval.nemo"
\end{lstlisting}

Listing~\ref{lst:kalenjin-train} reproduces the historical v1-v invocation rather than a corrected confirmatory recipe.
Its \texttt{validation.json} argument is a known protocol defect: that manifest contained \texttt{dev\_test}-origin rows also represented in the clean-v3 test set.
It must not be reused for future checkpoint selection; a corrected invocation should pass \texttt{validation\_dev.json}.
This correction does not retroactively make the reported v1-v selection or test result independent.

The action name retains the historical word \texttt{smoke}; at this stage it launches a full run and should not be interpreted as a tiny test.
For future maintenance, action names should be migrated to semantic names such as \texttt{train\_h200} while preserving backward-compatible aliases in experiment records.

\subsection{Foreground-Wait True-Streaming Evaluation}

Evaluation is shorter than training and benefits from a wait action that returns a structured report to the local client.
The \texttt{streaming\_eval\_h100\_wait} action calls the evaluator through \texttt{.remote()} rather than creating an unowned child with \texttt{.spawn()}.
The outer command may still use \texttt{-{}-detach}; Modal then owns the application while the entrypoint waits for its remote input.

\begin{lstlisting}[caption={True-streaming internal evaluation with a lifecycle-safe wait action.},label={lst:kalenjin-eval}]
MODAL_PROFILE=c-elo KALENJIN_INCLUDE_PREMIUM_GPUS=1 \
modal run --detach \
  training/nemotron-asr-kalenjin/modal_train.py \
  --action streaming_eval_h100_wait \
  --run-name \
    kalenjin_v1_v_h200_from_v1iv_full159h_cleanv3_lr5e7_70k_cleanval \
  --manifest-run-name kalenjin_full_clean_v3 \
  --split test \
  --model-nemo-path \
    "/data/checkpoints/kalenjin_v1_v_h200_from_v1iv_full159h_cleanv3_lr5e7_70k_cleanval/kalenjin_v1_v_h200_from_v1iv_full159h_cleanv3_lr5e7_70k_cleanval/2026-07-01_09-36-59/checkpoints/kalenjin_v1_v_h200_from_v1iv_full159h_cleanv3_lr5e7_70k_cleanval.nemo" \
  --target-lang sw-KE \
  --att-context-size "[56,13]" \
  --batch-size 8 \
  --eval-suffix full_cleanv3_test_v1v
\end{lstlisting}

For the Kalenjin v1-v evaluation, completion was established by the child function's terminal status and terminal evaluation log, which recorded row count, duration, WER, elapsed time, and configuration.
The unified report contract proposed below additionally requires model and manifest hashes and explicitly nullable CER fields.

\section{True-Streaming Evaluation Protocol}
\label{app:true-stream-protocol}

\subsection{Why Offline Transcription Is Not a Substitute}

An offline ASR evaluation typically presents the complete utterance to the encoder and decodes once.
That method is valuable for comparison but cannot validate a streaming state machine.
The deployed recognizer receives bounded chunks, carries acoustic and decoder state between calls, exposes partial text before the utterance ends, and must finalize without access to arbitrary future frames.
Consequently, the project evaluates the champion candidate with the same cache-aware mechanism used by the service.

\subsection{Evaluator Procedure}

The recorded evaluators share the streaming operations below, although report serialization varied across experiment stages.
Digest verification was enforced during champion transfer and preservation rather than uniformly inside every historical evaluator.

For a manifest $\mathcal{D}=\{(x_i,y_i,d_i)\}_{i=1}^{N}$, the evaluator performs the following operations:

\begin{enumerate}
   \item restore the selected \texttt{.nemo} archive and, for a preserved champion copy, verify its digest before evaluation;
  \item set the language prompt to \texttt{sw-KE};
  \item set the encoder attention context to \texttt{[56,13]};
  \item initialize the cache-aware streaming inferencer and greedy batched RNN-T decoder;
  \item load each waveform at the expected sampling rate;
  \item partition it into the same bounded chunks used by the NeMo streaming utility;
  \item feed chunks in chronological order while carrying encoder caches and decoder state;
  \item finalize one hypothesis only after the sample is complete;
  \item normalize reference and hypothesis with the frozen evaluation policy;
  \item accumulate edit counts globally rather than averaging per-utterance ratios;
  \item write aggregate metrics and available per-example outputs to the stage-specific report or terminal log.
\end{enumerate}

The batch is an efficiency device, not permission to mix states.
Each stream owns its cache and is removed from the active buffer when its waveform ends.
Variable-length batches must therefore preserve sample identity as streams enter and leave the buffer.

\subsection{Metric Definitions}

Let $S_w$, $D_w$, and $I_w$ be the corpus-level word substitutions, deletions, and insertions, and let $N_w$ be the number of words in all normalized references.
The word error rate is

\begin{equation}
\wer = \frac{S_w + D_w + I_w}{N_w}.
\label{eq:wer-corpus}
\end{equation}

Let $S_c$, $D_c$, $I_c$, and $N_c$ denote the analogous character counts.
Conventional character error rate is

\begin{equation}
\cer = \frac{S_c + D_c + I_c}{N_c}.
\label{eq:cer-corpus}
\end{equation}

Word-boundary inconsistency is common in the evaluated languages and can dominate character edits without implying a comparable acoustic failure.
For diagnostic use, spaces are removed after text normalization and before character alignment:

\begin{equation}
\nscer = \frac{S_{c,\neg\mathrm{space}} + D_{c,\neg\mathrm{space}} + I_{c,\neg\mathrm{space}}}
{N_{c,\neg\mathrm{space}}}.
\label{eq:nscer-corpus}
\end{equation}

NS-CER complements but does not replace WER.
It answers whether the character sequence is approximately correct when token boundaries vary; WER remains the primary lexical metric.
No result should be compared with an external score unless case-folding, punctuation, marker removal, Unicode normalization, and space treatment are identical.

\subsection{Latency and Throughput}

For total audio duration $T_a$ and wall-clock inference time $T_i$, the real-time factor is

\begin{equation}
\mathrm{RTF}=\frac{T_i}{T_a},
\qquad
\mathrm{RTFx}=\frac{T_a}{T_i}.
\label{eq:rtf}
\end{equation}

The reported Kalenjin v1-v clean-v3 adaptive diagnostic processed 5.33 hours in 92.16 seconds, corresponding to an aggregate RTF of approximately 0.0048, or about 208 times real time, for that batched H100 pass.
This is a throughput measurement, not an end-user latency measurement.
A complete product-level latency characterization would additionally require first-token latency, partial-update interval, endpointing delay, finalization delay, GPU cold start, network transport, and concurrent-stream behavior; those measurements were not collected in this study.

\subsection{Proposed Unified Evaluation Report Contract}

Historical evaluators archived stage-relevant metrics, but their schemas were not uniform and not every report included hashes or both CER variants.
For future frozen evaluations, we define the following minimum schema:

\begin{lstlisting}[caption={Proposed minimum schema for an immutable evaluation report.},label={lst:evaluation-schema}]
{
  "report_schema": "celo.asr.eval.v1",
  "language": "kalenjin",
  "split": "test",
  "label_version": "clean_v3",
  "manifest_sha256": "...",
  "model_sha256": "...",
  "prompt": "sw-KE",
  "attention_context": [56, 13],
  "decoder": "greedy_batched_rnnt",
  "streaming": true,
  "rows": 2411,
  "audio_hours": 5.33,
  "wer": 0.6874,
  "cer": null,
  "no_space_cer": null,
  "elapsed_seconds": 92.16,
  "normalization_policy": "kalenjin_eval_clean_v3"
}
\end{lstlisting}

This schema is a reproducibility recommendation introduced by this paper; it is not a claim that every historical report already conformed to it.
The v1-v evaluator emitted WER but not CER; the null fields explicitly mark those metrics as unreported rather than zero, and this paper does not reconstruct them from incomplete logs.

\section{Error Analysis Protocol}
\label{app:error-analysis}

Aggregate WER identifies whether a system improved, but not what should be changed next.
The project therefore combines edit-distance decomposition, transcript inspection, duration and domain stratification, and native-speaker review.

\subsection{Taxonomy}

\small
\begin{longtable}{L{2.8cm}L{5.2cm}L{5.9cm}}
\caption{Operational ASR error taxonomy used in this work.}\label{tab:error-taxonomy}\\
\toprule
Category & Observable signature & Likely response \\
\midrule
\endfirsthead
\toprule
Category & Observable signature & Likely response \\
\midrule
\endhead
Acoustic confusion & Phonetically nearby substitutions; errors grow with noise, channel distortion, or fast speech & Improve acoustic diversity, inspect waveform quality, and stratify by recorder/channel. \\
Lexical or orthographic confusion & Stable sounds but inconsistent spelling, apostrophes, diacritics, or word forms & Freeze a language-specific text policy with native-speaker review. \\
Boundary error & Correct-looking character stream split or joined into different words & Report WER with CER and NS-CER; avoid destructive whitespace normalization. \\
Code-switch error & Borrowed English or Swahili words are omitted, transliterated, or phonetically respelled & Preserve genuine spoken code-switching; remove only annotation wrappers such as \texttt{[cs]}. \\
Non-speech label learning & Output includes \texttt{cs}, \texttt{pause}, or \texttt{long pause} although those tokens were not spoken & Remove confirmed annotation markers in future cleaned label generations; retain a historical exception only when the policy is frozen, disclosed, and applied consistently. \\
Digit or identifier artifact & Dense digit strings or forms such as \texttt{we5107} create implausible lexical targets & Drop the full row when the intended spoken expansion cannot be recovered reliably. \\
Short-utterance over-generation & Hypothesis is much longer than a one- or two-word reference & Audit truncation and labels; add short examples only when trusted; test endpointing and insertion behavior. \\
Deletion dominance & Hypotheses are consistently shorter than references & Inspect low-volume speech, aggressive endpointing, duration truncation, and insufficient target-language adaptation. \\
Insertion dominance & Extra fluent-looking text appears after the reference content & Inspect RNN-T blank behavior, endpointing, non-speech tails, and mislabeled short clips. \\
Domain mismatch & Error clusters in medicine, agriculture, names, or public-service terminology & Add reviewed in-domain data or contextual biasing without contaminating the held-out split. \\
Prompt or language drift & Blank output or text from an unintended language & Verify the prompt route, initialization archive, tokenizer, and bridge lineage. \\
Streaming-state defect & Offline result is plausible but streaming result repeats, drops, or reorders text & Test cache ownership, state resets, chunk order, and end-of-stream flushing. \\
\bottomrule
\end{longtable}
\normalsize

\subsection{Edit Decomposition and Stratification}

Where evaluator outputs retained edit counts and metadata, project analysis decomposed errors and stratified them by available attributes.
A uniform future analysis should preserve insertion, deletion, and substitution counts separately and support the following slices:

\begin{itemize}
  \item reference word count and audio duration;
  \item scripted versus unscripted type;
  \item domain and recorder identity where metadata permits;
  \item presence of code-switched lexical material;
  \item duration quantile and transcript-length quantile;
  \item normalized character inventory;
  \item hypothesis/reference length ratio.
\end{itemize}

The highest-WER examples are useful but insufficient because long utterances naturally contain more opportunities for error.
The project review samples combined high-error examples, short-utterance failures, and examples selected for linguistic review.
A future structured review campaign should additionally include random and domain-balanced examples so that analysis is not driven only by worst cases.

\subsection{Targeted Native-Speaker Consultation and Proposed Review Protocol}

Targeted native-speaker consultation was conducted during development, but it was not a completed multi-annotator structured campaign.
For Kalenjin, speaker review confirmed that lexical material around code switches was valid while \texttt{[cs]}, \texttt{(cs)}, \texttt{[pause]}, and \texttt{(pause)} were annotation markers.
This finding directly motivated clean-v2.

For future review rounds, we propose a structured annotation task in which each selected row includes the audio, raw transcript, normalized transcript, model hypothesis, and the following response fields:

\begin{enumerate}
  \item Is the audio intelligible and primarily in the target language?
  \item Does the reference faithfully represent the spoken content?
  \item Is the normalization linguistically acceptable?
  \item Is apparent code-switching genuine speech or an annotation instruction?
  \item Are word boundaries and apostrophes acceptable variants?
  \item Should the row be retained, corrected, quarantined, or dropped?
  \item Which error category best explains the model output?
\end{enumerate}

Clean-v3 then dropped the entire row when the normalized reference contained the literal phrase \texttt{long pause}, any digit (including digit-only and alphanumeric forms), or fewer than three normalized tokens.
This was a deliberately conservative filter, not a claim that every number-bearing utterance is linguistically invalid.
The distinction matters: deleting only \texttt{we5107} would leave a grammatically broken target paired with unchanged audio, whereas dropping the complete row avoids creating new label noise.

\subsection{Short-Utterance Over-Generation Study}

Short references receive a dedicated analysis because one insertion can produce a large WER and because RNN-T endpointing errors are especially visible in this regime.
Define the length ratio

\begin{equation}
r_i = \frac{\max(1,|\hat{y}_i|_w)}{\max(1,|y_i|_w)},
\label{eq:length-ratio}
\end{equation}

where $|\cdot|_w$ counts words.
Clean-v3 excluded normalized references shorter than three tokens after over-generation was observed in this regime.
For subsequent error audits, we propose flagging rows with $|y_i|_w\leq 4$ and $r_i\geq 2$ for manual review; this threshold is an analysis rule, not an additional claim about the clean-v3 corpus filter.
The response is not to duplicate every short clip.
First determine whether the audio contains untranscribed speech, a long non-speech tail, clipping, or a faulty reference.
Only after data defects are excluded should decoding, blank calibration, chunk finalization, or targeted short-utterance sampling be changed.

\section{Operational Failure and Recovery Ledger}
\label{app:failure-ledger}

Infrastructure failures affected reproducibility and sometimes resembled model failures.
Table~\ref{tab:failure-ledger} records the principal incidents encountered during this project and the corrections adopted or recommended afterward.

\begin{longtable}{L{3.2cm}L{5.2cm}L{6.2cm}}
\caption{Failure modes, diagnoses, and durable corrections.}\label{tab:failure-ledger}\\
\toprule
Symptom & Diagnosis & Durable correction \\
\midrule
\endfirsthead
\toprule
Symptom & Diagnosis & Durable correction \\
\midrule
\endhead
Hugging Face \texttt{IncompleteRead} & Large Parquet/audio object transfer was interrupted & Use resumable library downloads and bounded retries; verify final object size and audio decoding. \\
Missing or undecodable waveform & Source row referenced an absent or malformed object & Log the source row and filename, increment \texttt{decode\_error}, skip deterministically, and preserve counts in the corpus ledger. \\
Replacement character in text & Earlier decoding introduced Unicode U+FFFD, making the intended grapheme unknowable & Drop the row rather than guessing the original character. \\
Tar-shard key or manifest mismatch & Archive member names and manifest paths were produced by inconsistent transformations & Build both from one structured record and verify every manifest key against the archive index before training. \\
Scheduler configuration error & Minimum learning rate was incompatible with the requested peak or inherited scheduler & Validate scheduler invariants before GPU allocation and log the resolved optimizer configuration. \\
GPU preemption or remote timeout & Cloud capacity or client connection ended while the job was active & Use persistent volumes, regular checkpoints, detached jobs, and resumable run lineage. \\
Long checkpoint-save stalls & Multi-gigabyte trainer checkpoints and NeMo exports blocked shutdown and validation completion & Treat saves as first-class timed phases; retain only policy-required checkpoints and allow sufficient function timeout/grace period. \\
``Runner shutting down too long'' after success & Modal cleanup exceeded its grace period after the final model and log were already committed & Judge success from training termination and durable artifact verification; record cleanup anomaly separately. \\
Child call immediately \texttt{Cancelled} & A local Modal entrypoint called \texttt{.spawn()} and exited, causing the unowned child input to be cancelled & Use \texttt{modal run -{}-detach} for long jobs and a wait action using \texttt{.remote()} when the entrypoint must own the result. \\
Local \texttt{Deadline exceeded} cancelled a healthy run & A foreground \texttt{.remote()} call remained coupled to a transient laptop/network connection & Prefer detached application ownership for multi-hour training; do not interrupt the client after submission. \\
Shell displays \texttt{quote>} & The command contains an unmatched single or double quote & Press Ctrl-C, reconstruct the command with one balanced quoted path, and use line continuations with no trailing characters after the backslash. \\
Volume CLI says path absent & Container path included the mount prefix \texttt{/data}, but volume browsing starts at the volume root & Remove \texttt{/data} when using \texttt{modal volume ls}; retain it inside functions. \\
Blank or wrong-language hypotheses & Raw multilingual initialization and prompt route were not sufficiently adapted & Establish a controlled bridge baseline; verify restored tokenizer, prompt, and checkpoint lineage. \\
Kikuyu v4 checkpoint average does not improve & The top-three v4 average scored 53.3887\% WER and 10.7683\% NS-CER versus 53.3505\% and 10.7256\% for the selected individual checkpoint; the available evidence does not isolate the cause & Treat averaged weights as a separate candidate and promote them only after measured validation and true-streaming evaluation. \\
\bottomrule
\end{longtable}

The failure ledger produced the following operating policy for subsequent runs.
First, a successful optimizer trajectory is not considered complete until the required archive and evaluation evidence are durable; the strengthened promotion contract also requires hashes.
Second, a cloud platform status is interpreted alongside the call graph: the local entrypoint and GPU child are distinct nodes.
Third, operational retries should not overwrite an existing run directory; retries receive an explicit attempt identifier or resume from a verified checkpoint.

\section{Deployment Architecture and Security Boundary}
\label{app:deployment}

\subsection{Isolated Language Services}

The two deployed champions in this study, Kikuyu and Dholuo, run as independent Modal applications and volumes.
This prevents a Dholuo deployment from replacing Kikuyu weights or changing its endpoint during testing.
Kalenjin remains a research candidate and is not represented as a live champion service.
The browser-facing language selector chooses a server-provided configuration; it does not receive a model path, Hugging Face token, or Modal secret.
The implemented configuration, direct streaming path, and trust boundary are summarized in Figure~\ref{fig:deployment}.

\subsection{What Authentication Does and Does Not Protect}

Route authentication prevents anonymous access to the application and configuration endpoint.
It does not make a WebSocket URL secret: an authenticated user can inspect any network request their browser makes.
Protection against unauthorized reuse therefore requires server-side enforcement at the ASR ingress, not obscurity in the JavaScript bundle.
The evaluated prototype validates the Supabase session on protected application and configuration routes, but that route-level check does not authorize the subsequent direct WebSocket hop.
A hardened release should validate a short-lived signed token or server-issued session credential when opening a stream, then enforce account and IP policies.
This stream-ingress token control was not implemented or evaluated in the present study.

Recommended controls include:

\begin{itemize}
  \item verify the Supabase session on every protected server route;
  \item mint short-lived, audience-bound stream tokens rather than exposing reusable service credentials;
  \item enforce allowed origins as a secondary browser control, never as the sole authorization mechanism;
  \item rate-limit stream creation per account and IP;
  \item meter audio seconds, concurrent streams, and request bursts server-side;
  \item cap session duration, utterance duration, and idle time;
  \item reject oversized or malformed frames before GPU processing;
  \item use TLS/WSS only and avoid sensitive query strings;
  \item log account, request identifier, duration, status, and aggregate usage without retaining raw audio by default;
  \item rotate secrets, separate development and production projects, and alert on abnormal use;
  \item keep private checkpoints in access-controlled volumes and private Hub repositories.
\end{itemize}

\subsection{Model Exfiltration Risk}

Normal inference does not transmit the \texttt{.nemo} archive to the browser; only audio, text hypotheses, and protocol metadata cross the network.
An attacker who reuses an unprotected inference endpoint can consume compute or query the model, but cannot directly download the checkpoint unless the storage or application exposes a file-serving path.
Model extraction through extensive black-box queries remains a broader research risk.
Authentication was implemented at the application and configuration-route layer; metering, anomaly detection, stream-token enforcement, legal controls, and extraction-specific output controls remain recommended hardening measures rather than evaluated protections in this study.

\section{Artifact Provenance and Preservation}
\label{app:provenance}

The champion package separates the deployment unit from its evidence.
This makes it possible to copy only the \texttt{.nemo} archive to a serving profile while retaining a complete preservation mirror elsewhere.

\subsection{Artifact Manifest}

\begin{lstlisting}[caption={Illustrative champion artifact manifest.},label={lst:artifact-manifest}]
{
  "schema": "celo.asr.champion.v1",
  "language": "dholuo",
  "model_family": "nemotron-3.5-asr-streaming-0.6b",
  "champion": "dholuo_v7_i_final_export",
  "evaluated_source_nemo": ".../checkpoints/dholuo_v7_i_h200_from_v6i_full207h_lr5e7_long40k_fullval.nemo",
  "preserved_best_validation_checkpoint": "...val_wer=0.3394-epoch=37.ckpt",
  "source_run": "2026-06-29_06-10-46",
  "source_checkpoint": "...val_wer=0.3394...ckpt",
  "nemo_path": "model/dholuo_v7_champion.nemo",
  "nemo_sha256": "...",
  "train_labels": "dholuo_full_clean_v1",
  "test_labels": "dholuo_full_clean_v1",
  "prompt": "sw-KE",
  "attention_context": [56, 13],
  "best_validation_wer": 0.33944,
  "evaluated_export_final_validation_wer": 0.33949,
  "test_wer": 0.339791,
  "test_cer": 0.095855,
  "test_no_space_cer": 0.081340,
  "test_rows": 5480,
  "private_hub_repo": "gateremark/dholuo-nemotron-3.5-asr-v1",
  "metric_provenance": "Test metrics apply to evaluated_source_nemo; the best-validation checkpoint was preserved separately."
}
\end{lstlisting}

The manifest itself is hashed, and the checksum file is generated after the directory is frozen.
When restoring into another Modal profile, the download helper computes the local \texttt{.nemo} SHA-256 digest and rejects a mismatch before placing the archive in the serving path.
The current helpers do not pin and independently verify a Hub commit revision, and the serving applications were not evaluated for startup digest attestation; both remain useful hardening steps.

\subsection{Preservation Tiers}

Three tiers balance recovery value against storage cost:

\begin{enumerate}
  \item \textbf{serving tier}: canonical \texttt{.nemo}, minimal model card, and digest;
  \item \textbf{champion evidence tier}: serving files plus evaluation outputs, manifest, resolved configuration, and provenance;
  \item \textbf{research resumption tier}: champion evidence plus selected trainer checkpoints and optimizer state needed to continue training.
\end{enumerate}

The Dholuo private repository is larger than the minimal Kikuyu mirror because it includes evaluation evidence and source checkpoints.
Both can run inference from the canonical \texttt{.nemo}; the additional files improve auditability and resumption, not recognition accuracy.

\section{Responsible Development, Data Governance, and Release}
\label{app:governance}

Low-resource language technology involves communities whose speech may be identifiable and whose linguistic conventions are not fully represented by standard tokenizers or externally imposed normalization.
The project therefore treats corpus access as permission to conduct the configured research, not as unrestricted permission to redistribute recordings, speaker metadata, or derived examples.

Redistribution of raw audio, speaker metadata, normalized manifests, and derived checkpoints is governed by the applicable source-dataset and base-model terms. This work therefore does not redistribute source-controlled audio or speaker metadata. Reproducibility is instead supported through processing scripts, checksums, aggregate corpus ledgers, source-row identifiers, resolved experiment configurations, and reconstruction instructions for authorized users.

The intended initial use is assisted transcription and controlled evaluation for Kikuyu and Dholuo speech.
The systems should not be represented as universally accurate across dialects, regions, age groups, recording environments, or specialized domains.
They are not suitable as the sole basis for legal, medical, financial, educational-assessment, or surveillance decisions.
Users should be informed that transcripts can contain substitutions, omissions, and insertions, and consequential use requires human verification.

Kalenjin is explicitly labeled research-in-progress because its aggregate WER remains high and dialect coverage requires further characterization.
Publishing the intermediate result is scientifically useful; deploying it without that qualification would be misleading.

\section{Extended Limitations and Threats to Validity}
\label{app:threats-validity}

\subsection{Internal Validity}

The continuation series changes both weights and, in some stages, corpus generations.
Improvements between clean-v1 and clean-v2 therefore combine further optimization with revised supervision unless the same checkpoint is re-evaluated on both label sets.
The Kalenjin v1-iii clean-v1/clean-v2 re-evaluation is included specifically to estimate the label-only component, but equivalent controls were not available for every transition.

Random seeds, sample order, GPU kernels, and resumed optimizer state are not fully replicated across multiple independent runs.
Reported deltas are therefore point estimates from an engineering lineage, not confidence intervals over stochastic retraining.
Future confirmatory work should rerun the final recipe with multiple seeds and report paired bootstrap intervals for held-out error rates.

\subsection{Construct Validity}

WER penalizes acceptable word-boundary variants and does not directly measure meaning preservation, named-entity accuracy, or user-perceived utility.
CER and NS-CER add complementary views but can make a semantically important one-character error appear small.
The project therefore avoids converting WER into an ``accuracy'' percentage and supplements metrics with native-speaker review and live testing.

The test manifests were held out from gradient updates, but repeated milestone evaluation informed subsequent decisions.
This creates adaptive pressure even without direct training on test audio.
A final publication claim should be confirmed on a newly frozen external or untouched set whose labels were not inspected during development.

\subsection{External Validity}

The corpora are Kenyan collections and may not cover every regional variety of Kikuyu, Dholuo, or Kalenjin.
Recorder and domain counts show useful diversity, but coverage is not a demographic representativeness guarantee.
Performance on clean single-speaker clips does not imply equal performance in meetings, telephony, broadcast audio, overlapping speech, music, or distant microphones.

The Swahili bridge is a successful practical choice in this project, not evidence that Swahili is always the best bridge for every African language.
Controlled comparison with the raw NVIDIA base, other regional checkpoints, self-supervised encoders, and matched-compute alternatives remains necessary.

\subsection{Comparison Validity}

Public PazaBench scores \cite{microsoft2026paza} and the internal scores in this paper do not share a test set or necessarily share normalization and decoding.
It is valid to use public results as context; it is not valid to order the systems in one leaderboard and claim state of the art without reproducing the same benchmark protocol.
Likewise, the strong batched H100 RTFx is not directly comparable with browser-to-server latency or an offline model measured on different hardware.

\section{Proposed Confirmatory Evaluation Protocol}
\label{app:confirmatory-study}

The engineering study supports a stronger follow-up experiment with a preregistered decision rule:

\begin{enumerate}
  \item freeze one final normalization policy per language;
  \item construct an untouched, speaker-disjoint external test set with documented dialect and domain coverage;
  \item obtain at least two independent native-speaker reviews for a stratified label sample;
  \item compare the raw NVIDIA base, Swahili bridge, final language model, Whisper \cite{radford2023whisper}, and public Paza baselines \cite{microsoft2026paza} on the same audio and normalization;
  \item evaluate both offline and true-streaming modes where the architecture permits;
  \item report WER, CER, NS-CER, insertion/deletion/substitution counts, RTF, first-token latency, finalization latency, and memory;
  \item compute paired bootstrap confidence intervals and test the prespecified primary contrast;
  \item publish the evaluator, manifest hashes, model revisions, and error-analysis protocol.
\end{enumerate}

This protocol defines the next stage of benchmark-comparable evaluation, including matched test conditions, native-speaker review, confidence intervals, and offline and streaming measurements.
It would also clarify whether the remaining Kalenjin error is primarily acoustic, orthographic, dialectal, or a consequence of initialization and decoder vocabulary.

\section*{Data and Artifact Availability}

The Kikuyu and Dholuo champion checkpoints are preserved in access-controlled Modal volumes and private Hugging Face repositories.
At the time of publication, the model weights and production deployment
artifacts are not publicly distributed.

The training and evaluation code is maintained by C-elo Labs. Requests for research access may be directed to the corresponding author and will be considered subject to the applicable source-dataset and base-model licenses, security review, and documentation readiness.

Raw ANV audio, speaker metadata, and other source-controlled materials are not redistributed by this work. Access to the original datasets must be obtained from the dataset provider under its applicable access and licensing terms.

\section*{Author Contributions}

Mark Gatere conceived the project, designed and implemented the data-processing, training, evaluation, preservation, and deployment pipelines, conducted the experiments, analyzed the results, and drafted the manuscript.

\section*{Ethics Statement}

This work develops speech-recognition systems for languages with limited representation in mainstream voice technology.
The potential benefit is improved access to digital services and language technology; the risks include transcription errors, uneven performance across speakers, unauthorized voice processing, surveillance misuse, and release of derived artifacts beyond source permissions.
The systems are evaluated as assistive recognizers, not as identity, intent, or truth detectors.

\section*{Acknowledgments}

The author thanks the creators and maintainers of NVIDIA NeMo and Nemotron 3.5 ASR, Tony Kipkemboi for the public Kenyan Swahili bridge checkpoint, the African Next Voices contributors and dataset maintainers, and the native speakers who reviewed language-specific transcript policies. The author thanks Eammon Kiprotich, Emmanuel Mwega and Zachary Odhiambo for native-speaker review, linguistic feedback, and assistance in validating transcription and text-normalization decisions.

No external monetary research funding was received. Modal provided in-kind startup compute credits that supported GPU-based model fine-tuning, evaluation, checkpoint preservation, and inference deployment.

\end{document}